\documentclass[acmtocl,acmnow]{acmtrans2m}

\usepackage{epsfig}
\usepackage{verbatim}
\usepackage{amsmath}
\usepackage{amssymb}
\usepackage{theorem}
\usepackage{graphicx}
\usepackage{color}
\usepackage{latexsym}
\usepackage{array} 

\newcommand{\K}{\mathord\mathbf{K}\,}
\newcommand{\nf}{\mathord\mathbf{not}\,}

\newcommand{\lfp}{\operatorname{\sf lfp}}

\newcommand{\com}[1]{}
\newcommand{\mif}{\leftarrow}

\newcommand{\subgc}{{\sc New Subgoal}}

\newcommand{\newsg}{{\sc New Subgoal}}
\newcommand{\pgmcr}{{\sc Program Clause Resolution}}
\newcommand{\ansret}{{\sc Positive Return}}
\newcommand{\anscr}{{\sc Positive Return}}

\newcommand{\negret}{{\sc Negative Return}}
\newcommand{\negsuc}{{\sc Negation Success}}
\newcommand{\negfail}{{\sc Negation Failure}}

\newcommand{\delay}{{\sc Delaying}}
\newcommand{\simpl}{{\sc Simplification}}
\newcommand{\compl}{{\sc Completion}}
\newcommand{\anscompl}{{\sc Answer Completion}}
\newcommand{\orcr}{{\sc Oracle Resolution}}

\newcommand{\cA}{{\cal A}}
\newcommand{\cC}{{\cal C}}
\newcommand{\cE}{{\cal E}}
\newcommand{\cF}{{\cal F}}
\newcommand{\cK}{{\cal K}}
\newcommand{\cO}{{\cal O}}
\newcommand{\cP}{{\cal P}}
\newcommand{\cS}{{\cal S}}

\newcommand{\cL}{{\cal L}}

\newtheorem{theorem}{Theorem}[section]

\newtheorem{corollary}[theorem]{Corollary}
\newtheorem{proposition}[theorem]{Proposition}
\newtheorem{lemma}[theorem]{Lemma}
\newdef{definition}[theorem]{Definition}
\newdef{remark}[theorem]{Remark}
\newdef{example}[theorem]{Example}

\newcommand\SLGJMT{{\bf SLG($\cO$)}}
\newcommand{\SLGO}{{\bf SLG($\cO$)}}
\newcommand\MKNF{{\em MKNF$_{WF}$}}

\newcommand\mycomment[1]{}

\title{Query-driven Procedures for Hybrid MKNF Knowledge Bases}

\author{JOS\'E J\'ULIO ALFERES \and MATTHIAS KNORR \and TERRANCE SWIFT \\CENTRIA, FCT, Universidade Nova de Lisboa}

\begin{abstract}
Hybrid MKNF knowledge bases are one of the most prominent tightly
integrated combinations of open-world ontology languages with
closed-world (non-monotonic) rule paradigms.  The definition of Hybrid
MKNF is parametric on the description logic (DL) underlying the ontology language, in the sense that
non-monotonic rules can extend any decidable DL language.
Two related semantics have been defined for Hybrid MKNF: one that is
based on the Stable Model Semantics for logic programs and one on 
the Well-Founded Semantics (WFS). Under WFS, the definition of 
Hybrid MKNF relies on a bottom-up computation that has polynomial 
data complexity whenever the DL language is tractable.
Here we define a general query-driven procedure for Hybrid MKNF that
is sound with respect to the stable model-based semantics, and sound
and complete with respect to its WFS variant.
This procedure is able to answer a slightly restricted form of
conjunctive queries, and is based on tabled rule evaluation extended
with an external oracle that captures reasoning within the ontology.
Such an (abstract) oracle receives as input a query along with
knowledge already derived, and replies with a (possibly empty) set of
atoms, defined in the rules, whose truth would suffice to prove the
initial query.  With appropriate assumptions on the complexity of the
abstract oracle, the general procedure maintains the data complexity
of the WFS for Hybrid MKNF knowledge bases.

To illustrate this approach, we provide a concrete oracle for
$\mathcal{EL}^{+}$, a fragment of the light-weight DL
$\mathcal{EL}^{++}$.  Such an oracle has practical use, as
$\mathcal{EL}^{++}$ is the language underlying OWL 2 EL, which is part 
of the W3C recommendations for the Semantic Web, and is tractable 
for reasoning tasks such as subsumption.
We show that query-driven Hybrid MKNF preserves polynomial data
complexity when using the $\mathcal{EL}^{+}$ oracle and WFS.

\end{abstract}






\begin{document}
            
\begin{bottomstuff} 
Author's addresses: J.J. Alferes, M. Knorr, and T. Swift: CENTRIA,
Faculdade de Ci\^{e}ncias e Tecnologia, Univ. Nova de Lisboa, 2825-516
Caparica, Portugal.  Preliminary versions of some of this material
appeared in \cite{AKS09} and \cite{KA:ECAI10}.
\end{bottomstuff}
            
\maketitle

\section{Introduction}\label{s:intro}

It is frequently claimed that integrating open world with closed world reasoning is a key issue for practical large-scale ontology applications.
As one example, \cite{PatelEtAl07} describes a large medical case study about matching patient records for clinical trials criteria containing up to millions of assertions.
In that clinical domain, open world reasoning is needed for radiology and laboratory data, because, for example, unless a
lab test asserts a negative finding, no arbitrary assumptions about the test can be made.
However, in pharmacy data, the closed world assumption can be used to infer that a patient is not on a specific medication unless it is asserted.

In general,  both ontologies and rules provide distinct strengths for the representation and interchange of knowledge in the Semantic Web and for applications of knowledge representation, such as the one described above. 
Expressive ontology languages are usually fragments of first-order logics represented in description logics (DLs) \cite{BCMNP07} and offer the deductive advantages of first-order logics with an open domain, while guaranteeing decidability.
Rules on the other hand  offer non-monotonic (closed-world) reasoning that can be useful for formalizing scenarios under (local) incomplete knowledge.
They also enable reasoning about fixed points (e.g., reachability), which cannot be expressed within first-order logic. 
Interest in ontologies, rules, and their combination is demonstrated by the development of ontology languages for the Semantic Web, such as OWL \cite{OWL_Primer}, and the growing interest on rule languages for the Semantic Web, cf.\ the RIF \cite{BK10:rif-overview} and the RuleML\footnote{\url{http://ruleml.org/}} initiatives. 

The majority of proposals for combining rules and ontologies (see,
e.g., related work in \cite{EILST08,KAH:AI11}) rely on one of the two
most common semantics for rules: the Well-Founded Semantics
(WFS) \cite{GRS91} or the Answer-Sets Semantics \cite{GL91}.  Both
semantics are widely used and allow closed-world reasoning and the
representation of fixed points.  Furthermore, the relationship between
the two semantics has been fully explored.  Of the two, the
Well-Founded Semantics is weaker (in the sense that it is more
skeptical w.r.t.\ derivable consequences), but it has the clear
advantage that its lower complexity is more suitable for applications
with large amounts of data, such as the medical case study described
above. 

Several formalisms have concerned themselves with combining decidable DLs with WFS rules~\cite{DM07,Luk10,EILS11,KAH:AI11}.  
Among these, the well-founded semantics for Hybrid MKNF knowledge bases (\MKNF{}), introduced first in \cite{KAH:ECAI08} and further refined in \cite{KAH:AI11}, is based on a three-valued extension of the logics of minimal knowledge and negation as failure (MKNF) \cite{Lif91}, and is the only one that allows knowledge about instances to be fully inter-definable between rules and an ontology that is taken as a parameter of the formalism.

\MKNF{} is defined using a monotonic fixpoint operator that computes in each iteration step, besides the usual immediate consequences from rules, the set of all atoms derivable from the ontology that is augmented with the already proven atomic knowledge.
The least fixpoint of the \MKNF{} operator coincides with the original
WFS~\cite{GRS91} if the ontology is empty, and coincides with the
semantics of the ontology if there are no rules; in addition, if the DL underlying the ontology language is polynomial, then \MKNF{} retains a polynomial data complexity.
Furthermore, \MKNF{} is sound with respect to the semantics of \cite{MR10} for MKNF knowledge bases (KBs), which is based on the Answer-Set Semantics and coincides with answer-sets of logic programs~\cite{GL91} if the ontology is empty.

In one sense, the fixpoint operator of \MKNF{} provides a way to compute, in a naive bottom-up fashion, all consequences of a knowledge base. 
However, such an approach is impractical for large knowledge bases.
Consider the medical case study above: knowledge of whether a specific patient is using a certain medication does not require knowledge of the medications of thousands of other patients.  Thus, despite its polynomial complexity, bottom-up computation of \MKNF{} does not scale to enterprise applications, much less to those of the Semantic Web.
A query-driven procedure corresponding to the semantics of \MKNF{} that only consults information relevant for a specific patient is clearly preferable.

This paper presents such a querying mechanism, called \SLGO{}, that is sound and complete for \MKNF{} \cite{KAH:AI11}, and sound for MKNF knowledge bases of \cite{MR10}.
\SLGO{} accepts DL-safe conjunctive queries, i.e., conjunctions of predicates with variables where queries have to be ground before being processed by the DL reasoner, returning all correct answer substitutions for variables in the query.
To the best of our knowledge, \SLGO{} is the first query-driven, top-down like procedure for knowledge bases that tightly combines an arbitrary decidable ontology language with non-monotonic rules.

\SLGO{} applies to any DL and under certain conditions maintains the data complexity of \MKNF{}.
To show that these conditions are realistic, we also provide a concrete oracle, with practical usage, namely for $\mathcal{EL}^{+}$.
$\mathcal{EL}^{+}$ is a fragment of the light-weight description logics $\mathcal{EL}^{++}$, which is the DL underlying OWL 2 EL -- one of the tractable profiles \cite{OWL2Pro} of OWL 2 -- and thus part of the W3C recommendations for the Semantic Web.
We show that the oracle thus defined is correct with respect to the general procedure and maintains the polynomial data complexity of \MKNF{} for such a polynomial DL.

\subsection*{The gist of the approach}\label{s:gist}

The main element of our approach addresses the interdependency of the ontology and rules.
In particular, \SLGO, presented in Section \ref{s:SLGO}, extends SLG resolution~\cite{CheW96}, which evaluates queries posed to normal logic programs, i.e., sets of non-disjunctive non-monotonic rules, under WFS.
SLG is a form of resolution that handles loops within the program, and does not change the data complexity of WFS.
It does that by resorting to already computed results in a forest of derivation trees, a technique also known as \emph{tabling}.

To adjoin an ontology to rules, the first thing that needs to be done is to allow an SLG evaluation to make calls to an inference engine for the ontology.
Since MKNF is parametric on any given decidable ontology formalism,\footnote{In fact, theoretically the limitation to decidable ontology formalisms is not strictly needed, but it is a pragmatic choice to achieve termination and complexity results in accordance  with decidable ontology languages, such as OWL \cite{OWL_Primer}.} the inference engine is viewed in \SLGO{} as an oracle.
In fact, every time \SLGO{} selects an atom,
the oracle's inference engine may be called, in case the atom is not provable by the rules alone.
Such a queried atom, say $\tt{P(a)}$, might thus be provable in the ontology but only if a certain set of atoms in turn is provable via rules.
Our approach captures this by allowing the oracle to return a new rule, say ${\tt P(a)} \mif Goals$, which has the property that a (possibly empty) set $Goals$, in addition to the axioms in the ontology and the atoms already derived from the combined knowledge base, would be sufficient to prove $\tt{P(a)}$.
\SLGO{} then treats these new rules just as if they were part of the knowledge base. 

Note that getting these conditional answers does not endanger decidability (or tractability, if it is the case) of reasoning in the ontology alone.
In fact, it is easy to conceive of a modification of a tableau-based
inference engine for an ontology that is capable of returning these
conditional answers and is decidable if the tableau algorithm is.
Simply add all the atoms that are defined in the rules to the
ontology, then proceed with the tableau as usual, but collect all those
added facts that have been used in the proof.
Under some assumptions on the complexity of the oracle, it is shown (in Section \ref{s:prop} along with some other properties) that \SLGO{} also retains the data complexity of \MKNF{}.

The second element of our approach arises from the need to properly combine the classical negation usually appearing in the ontology language with the non-monotonic negation of rules.
This problem, which is solved by the semantics of~\cite{KAH:AI11}, is similar to the issue of {\em coherence} that arises when adding classical (or strong) negation to logic programs \cite{GL91,pw90:neginfu,PA92}: the classical negation must imply negation by default.
In our case, if the ontology entails that some atom $A$ is false, then perforce the default negation $\nf A$ must hold as well.
The derivation must accordingly be modified since the proof of $\nf A$ cannot simply rely on the failure of the proof of $A$ as it is usual in Logic Programming.
For that purpose, an alternating fixpoint approach is used in the bottom-up construction defined in \cite{KAH:AI11}, where two alternating fixpoint operators are applied to two different sets of rules.
In each iteration step, the fixpoint construction alternates between deriving more true atoms, and more (default) false atoms; when deriving more true atoms, the original set of rules is used; when deriving more default false atoms, a transformed set of rules is used so that rules with head $A$ are removed if $\neg A$ holds.
This ensures that $\nf A$ is derived (see Section \ref{sec:mknfwf} for details).

Adapting the alternating fixpoint for the top-down derivation would
result in a procedure significantly different from the original SLG.
So instead, we transform the original knowledge base to ensure
coherence without the need for alternating between two sets of rules.
This approach is simpler to understand as it is a more direct
extension of SLG and separates the concerns of coherence from those of
top-down derivation; in addition the transformational approach should
also facilitate implementations of \SLGO{}.  Indeed, one can rely more
directly on the existing implementations that follow closely
SLG.\footnote{We have already made some experiments on the
implementation of \SLGO{} \cite{GAS10:padl}, relying on the XSB-Prolog
implementation.} Accordingly, Section~\ref{s:altcomp} defines the above
mentioned transformation of the knowledge base.  The transformation
itself provides an alternative formulation of \MKNF{} and is another
result of the paper.

Finally, in Section \ref{s:el+oracle}, we provide a concrete oracle for $\mathcal{EL}^{+}$.
Our approach includes a preprocessing step that applies the subsumption algorithm\footnote{Such as the one included in Pellet (\url{http://clarkparsia.com/pellet/}) or CEL (\url{http://lat.inf.tu-dresden.de/systems/cel/})}
for $\mathcal{EL}^{+}$ to compute all the subsumption relationships contained in the DL knowledge base and then remove redundant information with respect to answering queries.
The resulting reduced DL knowledge base is then translated into rules and can be directly combined with the set of rules contained in the combined knowledge base, so that \SLGO{} can be applied for querying.
This direct integration of the oracle into the querying mechanism, as we show, immediately ensures that the data complexity of \MKNF{} is maintained, i.e., the $\mathcal{EL}^{+}$ oracle is polynomial.

\section{Preliminaries}\label{s:prel}

We assume a basic understanding of the Well-Founded Semantics \cite{GRS91} and first-order logics, in particular notions related to Logic Programming and resolution (see e.g. \cite{Llo87}).
In this section we recall basic concepts that we rely on in the following sections.
In particular, we present description logics using the DL $\mathcal{ALC}$, the syntax of Hybrid MKNF knowledge bases, and their well-founded semantics.

\subsection{Description Logics} \label{sec:elprelim}

We recall general notions for description logics, basing our examples
on $\mathcal{ALC}$ with role inclusions although our work is in
principle applicable to any DL.  Afterwards, since we present a
concrete oracle for $\mathcal{EL}^{+}$ in Section~\ref{s:el+oracle},
we also review the syntax of that DL.  We refer to \cite{BCMNP07} for
a general and thorough overview of description logics

We start by recalling the syntax and semantics for a general DL $\mathcal{DL}$. 
DLs define \emph{concept descriptions} inductively with the help of a set of constructors, starting with a set $\mathsf{N_{C}}$ of \emph{concept names}, a set $\mathsf{N_{R}}$ of \emph{role names}, and a set $\mathsf{N_{I}}$ of \emph{individual names}.
Concept descriptions of $\mathcal{DL}$ are formed using a set of constructors, and the upper part of Table \ref{table1} shows the constructors of $\mathcal{ALC}$.
There, and in general, we use $a$ and $b$ to denote individual names, $R$ and $S$ to denote role names, and $C$ and $D$ to denote concept descriptions (all possibly with indices).

\begin{table}[t]
{\caption{Syntax and semantics of $\mathcal{ALC}$ with role inclusions.}\label{table1}}
\renewcommand{\arraystretch}{1.2}
\[\begin{array}{p{2.8cm}p{2.5cm}p{6.5cm}}
\hline \hline
Name & Syntax & Semantics\\
\hline
top & $\top$ & $\Delta^{\mathcal{I}}$ \\
bottom & $\perp$ & $\emptyset$ \\
negation & $\neg C$ & $\Delta^{\mathcal{I}}\setminus C^{\mathcal{I}}$ \\
conjunction & $C\sqcap D$ & $C^{\mathcal{I}} \cap D^{\mathcal{I}}$ \\
disjunction & $C\sqcup D$ & $C^{\mathcal{I}} \cup D^{\mathcal{I}}$ \\
existential restriction & $\exists R.C$ & $\{x\in \Delta^{\mathcal{I}} \mid \exists y\in \Delta^{\mathcal{I}} : (x,y)\in R^{\mathcal{I}} \wedge y\in C^{\mathcal{I}} \}$ \\
value restriction & $\forall R.C$ & $\{x\in \Delta^{\mathcal{I}} \mid \forall y\in \Delta^{\mathcal{I}} : (x,y)\in R^{\mathcal{I}}$ implies $y\in C^{\mathcal{I}} \}$ \\
\hline
GCI & $C\sqsubseteq D$ & $C^{\mathcal{I}} \subseteq D^{\mathcal{I}}$ \\
RI & $R_{1} \circ \cdots \circ R_{k} \sqsubseteq R$ &  $R_{1}^{\mathcal{I}} \circ \cdots \circ R_{k}^{\mathcal{I}} \sqsubseteq R^{\mathcal{I}}$\\
\hline
concept assertion & $C(a)$ & $a^{\mathcal{I}}\in C^{\mathcal{I}}$ \\
role assertion & $R(a,b)$ & $(a^{\mathcal{I}},b^{\mathcal{I}})\in R^\mathcal{I}$ \\
\hline \hline
\end{array}\]
\end{table}

The semantics of $\mathcal{DL}$-concept descriptions is defined in terms of an \emph{interpretation} $\mathcal{I} = (\Delta^{\mathcal{I}},\cdot^{\mathcal{I}})$.
The domain $\Delta^{\mathcal{I}}$ is a non-empty set of individuals and the \emph{interpretation function} $\cdot^{\mathcal{I}}$ maps each concept name $A\in \mathsf{N_{C}}$ to a subset $A^{\mathcal{I}}$ of $\Delta^{\mathcal{I}}$, each role name $R\in\mathsf{N_{R}}$ to a binary relation $R^{\mathcal{I}}$ on $\Delta^{\mathcal{I}}$, and each individual name $a\in\mathsf{N_{I}}$ to an individual $a^{\mathcal{I}}\in \Delta^{\mathcal{I}}$.
The extension of $\cdot^{\mathcal{I}}$ to arbitrary concept descriptions is inductively defined as shown in the third column of Table \ref{table1} for $\mathcal{ALC}$.

A $\mathcal{DL}$ \emph{TBox} $\mathcal{T}$ is a finite set of \emph{general concept inclusions (GCIs)} and possibly \emph{role inclusions (RIs)}, and both their syntax can be found in the middle of Table \ref{table1}.
An interpretation is a \emph{model} of a TBox $\mathcal{T}$ if, for each GCI and RI in $\mathcal{T}$, the conditions given in the third column of Table \ref{table1} are satisfied.
In the definition of the semantics of RIs, the symbol $'\circ'$ denotes composition of binary relations.

A $\mathcal{DL}$ \emph{ABox} $\mathcal{A}$ is a finite set of \emph{concept assertions} for concept descriptions $C$ and \emph{role assertions} for role names $R$ whose syntax can be found in the lower part of Table \ref{table1}.
ABoxes are used to describe a snapshot or state of the world.
An interpretation $\mathcal{I}$ is a \emph{model} of an ABox $\mathcal{A}$ if, for each concept assertion and role assertion in $\mathcal{A}$, the conditions given in the third column of Table \ref{table1} are satisfied.

A $\mathcal{DL}$ knowledge base $\cO$ consists of a $\mathcal{DL}$ TBox $\mathcal{T}$ and a $\mathcal{DL}$ ABox $\mathcal{A}$, and $\mathcal{I}$ is a \emph{model} of $\cO$ if it is a model of both $\mathcal{T}$ and $\mathcal{A}$.

One of the main inference problems in DLs, actually the one considered in \cite{BBL05} for $\mathcal{EL}^{++}$, is subsumption.
Given two $\mathcal{DL}$-concept descriptions $C$, $D$ we say that $C$ \emph{is subsumed by} $D$ w.r.t.\ the TBox $\mathcal{T}$ ($C \sqsubseteq_{\mathcal{T}} D$) iff $C^{\mathcal{I}} \subseteq D^{\mathcal{I}}$ for all models $\mathcal{I}$ of $\mathcal{T}$.
In addition, we recall the {\em instance problem} since, as we will see below, it is the reasoning  task we are interested in when answering top-down queries in our system combining rules and an oracle to an ontology.
An individual name $a$ is an \emph{instance of a concept} $C$ in ABox $\mathcal{A}$ w.r.t.\ a TBox $\mathcal{T}$ if $a^{\mathcal{I}}\in C^{\mathcal{I}}$ for every common model $\mathcal{I}$ of $\mathcal{A}$ and $\mathcal{T}$.
Definition~\ref{def:role-inst} extends the instance problem to instances of roles, a non-standard reasoning task.

\begin{definition} \label{def:role-inst}
A pair of individuals $(a,b)$ is an \emph{instance of a role} $R$ in ABox $\mathcal{A}$ w.r.t.\ a TBox $\mathcal{T}$ if $(a^{\mathcal{I}},b^{\mathcal{I}})\in R^{\mathcal{I}}$ for every common model $\mathcal{I}$ of $\mathcal{A}$ and $\mathcal{T}$.
\end{definition}
The above definition will be useful, since in the oracle we define queries for instances of roles, i.e., binary predicates, as well as concepts.

\subsubsection{$\mathcal{EL}^{+}$} 

The tractable DL $\mathcal{EL}^{+}$ is a large fragment\footnote{We omit nominals and concrete domains as concept constructors.} of the DL $\mathcal{EL}^{++}$ \cite{BBL05}.
It is obtained by restricting the allowed concept constructors to $\top$, $\bot$, $\neg$, $\sqcap$, and $\exists$, i.e., negations, disjunctions, and value restrictions are not allowed.
All the remaining notions, in particular the semantics, carries over from the general case.

Two remarks regarding the expressivity of $\mathcal{EL}^{+}$ are in order.
First, RIs allow expression of \emph{role hierarchies} $R\sqsubseteq S$, \emph{transitive roles}  using $R \circ R \sqsubseteq R$, \emph{right-identity rules} $R\circ S \sqsubseteq S$, and \emph{left-identity rules} $S\circ R\sqsubseteq S$.
Second, \emph{disjointness} of complex concept descriptions $C$, $D$ (and \emph{unsatisfiability} of a concept $C$), can be expressed by $C \sqcap D \sqsubseteq \perp$ (resp. $C \sqsubseteq \perp$).

\subsection{Syntax of Hybrid MKNF Knowledge Bases}

Hybrid MKNF knowledge bases, as introduced in \cite{MR10}, are essentially formulas in the logics of minimal knowledge and negation as failure \cite{Lif91}, i.e., first-order logics with equality and two modal operators, $\mathbf{K}$ and $\mathbf{not}$, allowing inspection of the knowledge base.
At an intuitive level, given a first-order formula $\varphi$, $\K\varphi$ asks whether $\varphi$ is known, i.e., true in all models of the related Hybrid MKNF knowledge base $\cK$, while $\nf \varphi$ is used to check whether $\varphi$ is not known.
Hybrid MKNF knowledge bases consist of two components, a decidable
DL knowledge base  translatable into a first-order logic, and a finite set of rules.

\begin{definition}\label{d:MKNFKB}
Let $\mathcal{O}$ be a DL knowledge base built over a language $\mathcal{L}$ with distinguished sets of countably infinitely many variables $\mathsf{N_{V}}$, and finitely many individuals $\mathsf{N_{I}}$ and predicates $\mathsf{N_{P}}$, where $\mathsf{N_C}\cup\mathsf{N_R}\subseteq \mathsf{N_P}$.
An atom $P(t_{1},\ldots,t_{n})$, where $P\in \mathsf{N_{P}}$ and $t_{i}\in \mathsf{N_{V}}\cup \mathsf{N_{I}}$,  is called a \emph{DL-atom} if $P$ occurs in $\mathcal{O}$,  otherwise it is called \emph{non-DL-atom}.
An MKNF rule $r$ has the following form where, for all $i$ and $j$, $H$, $A_{i}$, and $B_{j}$ are atoms: 
\begin{equation}
H\leftarrow A_{1},\ldots, A_{n}, \nf B_{1},\ldots,\nf B_{m}.
\end{equation}
$H$ is called the \emph{(rule) head}, and the sets $\{A_{1},\ldots,A_n\}$ and $\{\nf B_{1},\ldots \nf B_m\}$ form the \emph{body} of the rule.
\emph{Literals} are \emph{positive literals} $A$ or \emph{negative literals} $\nf B$.
We abbreviate rules by $H\leftarrow \mathcal{B}$, splitting $\mathcal{B}$ into two sets $\mathcal{B}^+$ (positive literals) and $\mathcal{B}^-$ (negative literals).
A rule $r$ is \emph{positive} if $m=0$ and a \emph{fact} if $n=m=0$. 
A \emph{program} $\mathcal{P}$ is a finite set of MKNF rules and called \emph{positive} if all its rules are positive.
A  \emph{Hybrid MKNF knowledge base} $\mathcal{K}$ is a pair $(\mathcal{O},\mathcal{P})$.
 The \emph{ground instantiation} of $\mathcal{K}$ is the KB $\mathcal{K}_{G}=(\mathcal{O},\mathcal{P}_{G})$ where $\mathcal{P}_{G}$ is obtained from $\mathcal{P}$ by replacing each rule $r$ of $\mathcal{P}$ with a set of rules substituting each variable in $r$ with constants from $\mathcal{K}$ in all possible ways.
\end{definition}

In this definition and in the rest of the paper, we omit the modal operator $\mathbf{K}$ in rule heads and bodies.\footnote{The MKNF semantics in \cite{MR10} and \cite{KAH:AI11} requires the presence of these modal operators to ensure that the interaction between the DL KB $\mathcal{O}$ and the rules is limited to information that is known to hold. For our purposes, a simpler representation of models suffices, thus allowing us to simplify notation here.}

To ensure decidability, DL-safety is applied \cite{MR10,KAH:AI11}.
Intuitively, DL-safety constrains the use of rules to individuals actually appearing in the knowledge base under
consideration.
Since, as indicated in the introduction, we are especially interested in querying the knowledge base, care also must be taken to impose DL-safety on (conjunctive) queries:
\begin{definition}\label{d:dlsafecq}
An MKNF rule $r$ is \emph{DL-safe} if every variable in $r$ occurs in at least one (positive) non-DL-atom in the body of $r$.
A Hybrid MKNF knowledge base $\mathcal{K}$ is \emph{DL-safe} if all its rules are DL-safe.

A \emph{DL-safe conjunctive query} $q$ is a non-empty set, i.e., conjunction, of literals where each variable in $q$ occurs in at least one (positive) non-DL-atom in $q$. We also write $q$ as a rule $q(X_{i})\leftarrow A_{1}, \ldots, A_{n}, \nf B_{1}, \ldots, \nf B_{m}$ where $X_{i}$ is the (possibly empty) set of variables, appearing in the body.
\end{definition}
This restriction of conjunctive queries to DL-safety is not always necessary: for DLs like SHIQ, conjunctive query answering is decidable \cite{GLHS08} and we may make use of existing algorithms. However, for DLs where there is no known algorithm for conjunctive query answering or where the problem is not decidable, such as full $\mathcal{EL}^{++}$ \cite{Rosa07c}, the limitation is required to achieve decidability in Hybrid MKNF knowledge bases.
For simplicity of presentation, we impose the restriction throughout the paper.

\begin{example}\label{exPrel1}
We present a small technical example to illustrate the notions introduced in this section.
Consider the Hybrid MKNF knowledge base $\cK$ consisting of an $\mathcal{EL}^+$ KB $\mathcal{O}$ containing two TBox statements and one assertion and a set of MKNF rules.
Here and in the following examples we follow the convention that DL-atoms are capitalized, while non-DL-atoms start with lower case letters. 
\begin{align*}
\tt{C} & \sqsubseteq \tt{D} & \tt{C}(\tt{b}) &\\
\tt{C}\sqcap \tt{E} & \sqsubseteq \bot & &\\
\tt{p}(\tt{x}) & \leftarrow \nf \tt{D}(\tt{x}), \tt{o}(\tt{x}) & \tt{o}(\tt{a}) &\leftarrow\\
\tt{E}(\tt{x}) & \leftarrow \nf \tt{E}(\tt{x}), \tt{o}(\tt{x}) & \tt{o}(\tt{b}) &\leftarrow
\end{align*}
The ground instantiation $\cK_G$ is obtained by grounding both rules with $\tt{a}$ and $\tt{b}$.
Note that the atom $\tt{o}(\tt{x})$ ensures that both (non-ground) rules are DL-safe.
\end{example}

\subsection{Well-founded Semantics of Hybrid MKNF Knowledge Bases}\label{sec:mknfwf}

In this section, we recall the computation of the well-founded MKNF model from \cite{KAH:AI11}.\footnote{The well-founded MKNF semantics including the well-founded MKNF model, as presented in \cite{KAH:AI11}, is based on a complete three-valued extension of the original MKNF semantics of \cite{MR10}. In it, a model consists of two sets of sets of first-order interpretations; a 3-valued truth valuation is defined that exactly determines the semantics, and in which any MNKF formula can be evaluated. However, as here we are only interested in queries that are (conjunctions of) atoms,  we limit ourselves to the computation of the literals that are true and false. This is called the well-founded partition in \cite{KAH:AI11} but we term it the well-founded MKNF model here.}
We adopt that terminology here and recall the  notions relevant for its definition.
First, we present some notions from \cite{KAH:AI11} that are useful in the definition of the operators for obtaining that well-founded MKNF model. 

\begin{definition}\label{d:katoms}
Let $\mathcal{K}=(\mathcal{O},\mathcal{P})$ be a ground Hybrid MKNF knowledge base.
The set of known atoms of $\mathcal{K}$, ${\sf KA}(\mathcal{K})$, is the smallest set that contains (i) all positive literals occurring in $\mathcal{P}$, and (ii) a positive literal $\xi$ for each negative literal $\nf \xi$ occurring in $\mathcal{P}$.
For a subset $S$ of ${\sf KA}(\mathcal{K})$, the \emph{objective knowledge} of $S$ w.r.t.\ $\mathcal{K}$ is the set of first-order formulas ${\sf OB}_{\mathcal{O},S} = \{\pi(\mathcal{O})\}\cup \{\xi\mid (\mathbf{K})\,\xi\in S\}$ where $\pi(\mathcal{O})$ is the first-order translation of $\mathcal{O}$.
\end{definition}
Basically all literals appearing in the rules are collected in the set ${\sf KA}(\mathcal{K})$ as a set of positive literals while the objective knowledge ${\sf OB}_{\mathcal{O},S}$ provides a first-order representation of $\mathcal{O}$ together with a set of known/derived facts without the implicit modal operator $\mathbf{K}$.
For the computation of the three-valued MKNF model, the set ${\sf KA}(\mathcal{K})$ can be divided into true, undefined, and false literals.

\begin{example}
Recall $\cK$ from Example~\ref{exPrel1} and its ground instantiation $\cK_G$.
Then ${\sf KA}(\mathcal{K}_G)=\{\tt{p}(\tt{a}),\tt{p}(\tt{b}),\tt{D}(\tt{a}),\tt{D}(\tt{b}),\tt{E}(\tt{a}),\tt{E}(\tt{b}),\tt{o}(\tt{a}),\tt{o}(\tt{b})\}$.
\end{example}

We continue by defining an operator $T_{\mathcal{K}}$ that allows us to draw conclusions from positive Hybrid MKNF knowledge bases.

\begin{definition}\label{d:opRkDkTk}
Let $\mathcal{K}=(\mathcal{O},\mathcal{P})$ be a positive, ground Hybrid MKNF knowledge base.
The operators $R_{\mathcal{K}}$, $D_{\mathcal{K}}$, and $T_{\mathcal{K}}$ are defined on subsets of ${\sf KA}(\mathcal{K})$:
\begin{align*}
R_{\mathcal{K}}(S) =& \{H \mid  \mathcal{P} \text{ contains a rule of the form } H\leftarrow A_1,\ldots A_n \\ 
& \text{ such that, for all i, }1\leq i\leq n, A_{i}\in S\}\\
D_{\mathcal{K}}(S) =& \{\xi \mid \xi\in {\sf KA}(\mathcal{K}) \text{ and } {\sf OB}_{\mathcal{O},S}\models \xi\} \\
T_{\mathcal{K}}(S) =& R_{\mathcal{K}}(S)\cup D_{\mathcal{K}}(S)
\end{align*}
\end{definition}
$R_{\mathcal{K}}$ derives consequences from the rules while $D_{\mathcal{K}}$ obtains knowledge from the ontology $\mathcal{O}$ together with the information in $S$.

The operator $T_{\mathcal{K}}$ is shown to be monotonic in \cite{KAH:AI11}. So, by the Knaster-Tarski theorem \cite{Tar55}, it has a unique least fixpoint, denoted $\lfp(T_{\mathcal{K}})$, which is reached after a finite number of iteration steps (since the ground knowledge base $\cK$ is always finite).

The computation of the well-founded MKNF model follows the alternating fixpoint construction \cite{AvG89} of the well-founded semantics for logic programs.
This construction requires a reduction that turns a Hybrid MKNF knowledge base into a positive one to make $T_{\mathcal{K}}$ applicable.
\begin{definition}\label{d:MKNFtransformK}
Let $\mathcal{K}=(\mathcal{O},\mathcal{P})$ be a ground Hybrid MKNF knowledge base and $S\subseteq{\sf KA}(\mathcal{K})$.
The \emph{MKNF transform} $\mathcal{K}/S$ is defined as $\mathcal{K}/S=(\mathcal{O},\mathcal{P}/S)$, where
$\mathcal{P}/S$ contains all rules $H\leftarrow A_{1},\ldots, A_{n}$
for which there exists a rule 
\[
H\leftarrow A_{1}, \ldots, A_{n}, \nf B_{1},\ldots, \nf B_{m}
\]
in $\mathcal{P}$ with $B_{j}\not\in S$ for all $1\leq j\leq m$.
\end{definition}

\begin{example}
Consider again $\cK$ from Example \ref{exPrel1} and let $S$ be ${\sf KA}(\mathcal{K}_G)$.
Then $\mathcal{K}_G/S$ is obtained as follows:
\begin{align*}
\tt{C} & \sqsubseteq \tt{D} & \tt{C}(\tt{b}) &\\
\tt{C}\sqcap \tt{E} & \sqsubseteq \bot & &\\
\tt{o}(\tt{a}) &\leftarrow &  \tt{o}(\tt{b}) &\leftarrow
\end{align*}
The resulting KB is positive and we may apply $T_{\cK}$ and obtain $\{\tt{D}(\tt{b}),\tt{o}(\tt{a}),\tt{o}(\tt{b})\}$.
Note that $\tt{C}(\tt{b})$ is not explicitly mentioned since it does not occur in ${\sf KA}(\mathcal{K}_G)$.
It is nevertheless derivable from $\cK_G$.
\end{example}

The MKNF transform resembles the well-known answer-set transformation \cite{GL91} for logic programs.
Based on it, an antitonic operator can be defined, but it is shown
in \cite{KAH:AI11} that such an operator alone would not properly
treat a problem called {\em coherence}, i.e., classical negation would not enforce default negation.
Therefore, a second, slightly different transformation is introduced in \cite{KAH:AI11}.

\begin{definition}\label{d:MKNF-cohtransformK}
Let $\mathcal{K}=(\mathcal{O},\mathcal{P})$ be a ground Hybrid MKNF knowledge base and $S\subseteq{\sf KA}(\mathcal{K})$.
The \emph{MKNF-coherent transform} $\mathcal{K}//S$ is defined as $\mathcal{K}//S=(\mathcal{O},\mathcal{P}//S)$, where
$\mathcal{P}//S$ contains all rules $H\leftarrow A_{1},\ldots, A_{n}$ for which there exists a rule 
\[
H\leftarrow A_{1}, \ldots, A_{n}, \nf B_{1},\ldots, \nf B_{m}
\]
in $\mathcal{P}$ with $B_{j}\not\in S$ for all $1\leq j\leq m$ and ${\sf OB}_{\mathcal{O},S}\not\models \neg H$.
\end{definition} 

\begin{example}
Consider again $\cK$ from Example \ref{exPrel1} and let $S$ be $\emptyset$.
Then $\mathcal{K}_G//S$ is obtained:
\begin{align*}
\tt{C} & \sqsubseteq \tt{D} & \tt{C}(\tt{b}) &\\
\tt{C}\sqcap \tt{E} & \sqsubseteq \bot & \tt{p}(\tt{a}) & \leftarrow \tt{o}(\tt{a})\\
\tt{p}(\tt{b}) & \leftarrow \tt{o}(\tt{b}) & \tt{o}(\tt{a}) &\leftarrow\\
\tt{E}(\tt{a}) & \leftarrow \tt{o}(\tt{a}) & \tt{o}(\tt{b}) &\leftarrow
\end{align*}
The only rule that is removed is the one with head $\tt{E}(\tt{b})$, since $\tt{C}(\tt{b})$ holds and $\tt{C}$ and $\tt{E}$ are disjoint.
Hence, ${\sf OB}_{\mathcal{O},\emptyset}\models \neg \tt{E}(\tt{b})$. 
We can apply $T_{\cK_G}$ to the resulting positive KB and obtain ${\sf KA}(\mathcal{K}_G)\setminus \{\tt{E}(\tt{b}),\tt{D}(\tt{a})\}$.
\end{example}

Note the difference between Definitions \ref{d:MKNFtransformK} and \ref{d:MKNF-cohtransformK}: we also remove a rule from the MKNF-coherent transform, in case the classical negation of the head is derivable from the ontology augmented by $S$.

These two transformations can now be used to define two operators for Hybrid MKNF KBs as already hinted in the examples.
\begin{definition}\label{d:GammaK}
Let $\mathcal{K}=(\mathcal{O},\mathcal{P})$ be a ground Hybrid MKNF knowledge base and $S\subseteq
{\sf KA}(\mathcal{K})$.
We define:
\[
\Gamma_{\mathcal{K}}(S) = \lfp(T_{\mathcal{K}/S})\qquad \text{and} \qquad \Gamma'_{\mathcal{K}}(S) = \lfp(T_{\mathcal{K}//S}).
\]
\end{definition}

Both operators are shown to be antitonic \cite{KAH:AI11} and form the basis for defining the well-founded MKNF model.
Here we present its alternating computation.

\begin{definition}\label{d:ite}
Let $\mathcal{K}$ be a ground Hybrid MKNF knowledge base.
We define:
\begin{align*}
\mathbf{P}_{0} & = \emptyset & \mathbf{N}_{0} & = {\sf KA}(\mathcal{K})\\
\mathbf{P}_{n+1} & = \Gamma_{\mathcal{K}}(\mathbf{N}_{n}) & \mathbf{N}_{n+1} & = \Gamma'_{\mathcal{K}}(\mathbf{P}_{n}) \\
\mathbf{P}_{\omega} & = \bigcup\mathbf{P}_{i} & \mathbf{N}_{\omega} & = \bigcap\mathbf{N}_{i}
\end{align*}
\end{definition}
$\mathbf{P}_{\omega}$ contains everything that is necessarily true, while $\mathbf{N}_{\omega}$ contains everything that is not false.
Note that, by finiteness of the ground knowledge base, the iteration stops before reaching $\omega$.
It was shown in \cite{KAH:AI11} that the sequences are monotonically increasing, decreasing respectively.
The two fixpoints can also be used to detect whether a knowledge base is MKNF-consistent or not \cite{KAH:AI11}.

\begin{theorem}\label{t:consKB}
Let $\mathcal{K}=(\mathcal{O},\mathcal{P})$ be a ground Hybrid MKNF knowledge base.
$\mathcal{K}$ is MKNF-inconsistent iff $\Gamma_{\mathcal{K}}'(\mathbf{P}_{\omega})\subset \Gamma_{\mathcal{K}}(\mathbf{P}_{\omega})$ or $\Gamma'_{\mathcal{K}}(\mathbf{N}_{\omega})\subset \Gamma_{\mathcal{K}}(\mathbf{N}_{\omega})$ or $\mathcal{O}$ is inconsistent.
\end{theorem}
\noindent
Intuitively, MKNF-consistency requires that $\cO$ is (first-order)
consistent and that neither of the two additional conditions of
Theorem~\ref{t:consKB} succeed.\footnote{The formal definition of
MKNF-consistency from \cite{KAH:AI11} would require the complete
material on three-valued MKNF semantics, which we want to avoid.}
These two comparisons ensure that the two fixpoints never contain
contradictions and that there is no rule such that the truth value of
the (conjunction in the) body is greater than the truth value of the
head.

\begin{example}
Consider the following KB $\cK_1$:
\begin{align*}
\tt{P(a)} & \leftarrow \nf \tt{P(a)}& \tt{Q(a)} &\leftarrow  & \tt{Q} &\sqsubseteq \neg \tt{P}
\end{align*}
$\cK_1$ is MKNF-inconsistent, since $\tt{P(a)}$ is necessarily false, so the first rule violates the intuitive condition that the body should not have a higher truth value than the head.
In fact, the computation yields that $\tt{P(a)}$ is true and false at the same time, and the test reveals that.
Consider the KB $\cK_2$:
\begin{align*}
\tt{R}  \sqsubseteq &\, \neg \tt{P} & \,\tt{R}(\tt{a})  \\
\tt{P}(\tt{a})  \leftarrow &\, \nf \tt{u} & \tt{u}  \leftarrow & \,\nf \tt{u} 
\end{align*}
$\tt{P(a)}$ is false, but $\tt{u}$ is undefined, so that we obtain a rule with false head and undefined body, and this is detected with the test.
\end{example}

If $\mathcal{K}$ is MKNF-consistent, then the well-founded MKNF model exists.
\begin{definition}\label{d:MKNFWFS}
The \emph{well-founded MKNF model} $M_{WF}$ of an MKNF-consistent, ground Hybrid MKNF knowledge base $\mathcal{K}=(\mathcal{O},\mathcal{P})$ is defined as 
\[
M_{WF} = \{A\mid A\in \mathbf{P}_{\omega}\}\cup \{\pi(\mathcal{O})\} \cup \{\nf B \mid B\in {\sf KA}(\mathcal{K}) \setminus\mathbf{N}_{\omega}\}.
\]
\end{definition}

\begin{example}\label{exPrel2}
Consider again the running Example \ref{exPrel1}.
We provide the results of the computation.
\begin{align*}
\mathbf{P}_{0} & = \emptyset & \mathbf{N}_{0} & = {\sf KA}(\mathcal{K}_G)\\
\mathbf{P}_{1} & =  \{\tt{D}(\tt{b}),\tt{o}(\tt{a}),\tt{o}(\tt{b})\} & \mathbf{N}_{1} & = {\sf KA}(\mathcal{K}_G)\setminus \{\tt{E}(\tt{b}),\tt{D}(\tt{a})\}\\
\mathbf{P}_{2} & =  \{\tt{D}(\tt{b}),\tt{o}(\tt{a}),\tt{o}(\tt{b}),\tt{p}(\tt{a})\} & \mathbf{N}_{2} & = {\sf KA}(\mathcal{K}_G)\setminus \{\tt{E}(\tt{b}),\tt{D}(\tt{a}),\tt{p}(\tt{b})\}\\
\mathbf{P}_{3} & = \mathbf{P}_{2} & \mathbf{N}_{3} & = \mathbf{N}_{2}
\end{align*}
Thus we obtain the well-founded MKNF model as
\[
M_{WF} = \{\pi(\mathcal{O})\} \cup \{\tt{D}(\tt{b}),\tt{o}(\tt{a}),\tt{o}(\tt{b}),\tt{p}(\tt{a})\} \cup \{\nf \tt{E}(\tt{b}),\nf \tt{D}(\tt{a}),\nf \tt{p}(\tt{b})\} .
\]
Consequently, $\tt{E}(\tt{a})$ is undefined.
\end{example}

To ease some proofs in the following section, we also adapt the notion of unfounded sets \cite{GRS91} for Hybrid MKNF which relates to the sequence $\mathbf{N}_i$ (see \cite{KAH:AI11}).
The essential advantage is that the reasons why a certain atom is considered false are better characterized.
For that purpose, we first need to define a notion of dependency that captures more precisely the derivations from ${\sf OB}_{\mathcal{O},S}$, for some $S$, by the operator $D_{\mathcal{K}}$.

\begin{definition}\label{d:depOB}
Let $\mathcal{K}=(\cO,\mathcal{P})$ be a ground Hybrid MKNF knowledge base, $H$ an atom with $H\in {\sf KA}(\mathcal{K})$, and $S$ a (possibly empty) set of atoms with $S\subseteq {\sf KA}(\mathcal{K})$.
We say that $H$ \emph{depends} on $S$ if and only if
\begin{itemize}
\item[(i)] ${\sf OB}_{\mathcal{O},S}\models H$ and
\item[(ii)] there is no $S'$ with $S'\subset S$ such that ${\sf OB}_{\mathcal{O},S'}\models H$.
\end{itemize}
\end{definition}
Intuitively, $S$ is a minimal set that, in combination with $\cO$, allows us to derive $H$.
Note that there may exist several such minimal sets.
Based on this notion of dependency, the notion of an unfounded set can be extended to Hybrid MKNF KBs.

\begin{definition}\label{d:unfMKNF}
Let $\cK$ be a ground Hybrid MKNF knowledge base and $(T,F)$ a pair of sets of atoms with $T,F\subseteq {\sf KA}(\mathcal{K})$.
We say that $U\subseteq {\sf KA}(\mathcal{K})$ is an \emph{unfounded set}\index{unfounded set, Hybrid MKNF} (of $\cK$) \emph{with respect to $(T,F)$} if, for each atom $H\in U$, the following conditions are satisfied: 
\begin{itemize}
\item[(Ui)] for each rule $H \leftarrow \mathcal{B}$ in $\mathcal{P}$ at least one of the following holds.
\begin{itemize}
\item[(Uia)] Some atom $A$ appears in $\mathcal{B}$ and in $U\cup F$. 
\item[(Uib)] Some negative literal $\nf B$ appears in $\mathcal{B}$ and in $T$.
\item[(Uic)] ${\sf OB}_{\mathcal{O},T}\models \neg H$
\end{itemize} 
\item[(Uii)] for each (possibly empty) $S$ on which $H$ depends, with $S\subseteq {\sf KA}(\mathcal{K})$ and ${\sf OB}_{\mathcal{O},S}$ consistent, there is at least one atom $A$ such that ${\sf OB}_{\mathcal{O},S\setminus \{ A\}}\not\models H$ and $A$ in $U\cup F$.
\end{itemize}
The union of all unfounded sets of $\cK$ w.r.t.\ $(T,F)$ is called the \emph{greatest unfounded set} of $\cK$ w.r.t.\ $(T,F)$ and denoted $U_\cK(T,F)$.
\end{definition}

It can be shown that the computation of $\mathbf{N}_i$ based on $\Gamma'_\cK$ directly corresponds to the computation of the greatest unfounded set w.r.t.\ $(\mathbf{P}_{i-1},\mathbf{N}_{i-1})$.

\begin{example}
Consider the computation in Example~\ref{exPrel2}.  All three atoms
that were removed in the sequence of $\mathbf{N}_i$ ---
$\tt{E}(\tt{b})$, $\tt{D}(\tt{a})$, $\tt{p}(\tt{b})$ --- obviously
satisfy (Uii).  However these removed atoms satisfy different
conditions of (Ui).  For $\tt{E}(\tt{b})$, (Uic) applies.  In the case
of $\tt{D}(\tt{a})$, there is no rule with this head, so (Ui) is
vacuously true.  Finally, for $\tt{p}(\tt{b})$, (Uib) applies because
$\nf \tt{D}(\tt{b})$ occurs in the single rule for this atom, while
{\tt D(b)} is true.
\end{example}

\section{Alternative Computation of \MKNF{}}\label{s:altcomp}

As presented in Section \ref{s:prel}, the bottom-up computation of the well-founded MKNF model requires essentially two operators each with its own transformation of the knowledge base.
Using the operators directly would make the top-down procedure quite different from the original SLG procedure, which operates on a single logic program, and does not differentiate between the two phases of the alternating fixpoint.
To approximate the bottom-up computation to the SLG procedure, in this section we define that computation in a different way.
Namely, we transform the original knowledge base, by doubling the rules and the ontology in $\mathcal{K}$ using new predicates, and transform both so that a single operator and copy of the KB can be used.
As we shall see, a simpler bottom-up computation, with a single operator, performed over this single transformed knowledge base yields the same results as the one defined in  Section \ref{s:prel}, and in particular still guarantees that classical negation enforces default negation.

The first definition introduces two new special predicates for each predicate appearing in $\mathcal{K}$ based on which the transformation that doubles a knowledge base $\cK$ is defined.

\begin{definition}\label{d:doubling}
Let $\mathcal{K}=(\mathcal{O},\mathcal{P})$ be a Hybrid MKNF knowledge base.
We introduce new predicates $A^{d}$ and $NA$ for each predicate $A$ appearing in $\mathcal{K}$, and then constructively define
\begin{enumerate} 
\item $\cO^d$ by substituting each predicate $A$ in $\cO$ by $A^d$; and 
\item $\cP^d$ by transforming each rule 
\[H(\vec{t_H})\leftarrow A_{1}(\vec{t_{A1}}),\ldots, A_{n}(\vec{t_{An}}), \mathord{\nf} B_{1}(\vec{t_{B1}}),\ldots,\mathord{\nf} B_{m}(\vec{t_{Bm}})\] occurring in $\mathcal{P}$ into two rules:

\begin{itemize}
\item[(2a)] $H(\vec{t_H})\leftarrow A_{1}(\vec{t_{A1}}), \mathord{\nf} B_{1}^{d}(\vec{t_{B1}}),\ldots,\mathord{\nf} B_{m}^{d}(\vec{t_{Bm}})$ and either
\item[(2b.i)] $H^{d}(\vec{t_H})\leftarrow A_{1}^{d}(\vec{t_{A1}}),\ldots, A_{n}^{d}(\vec{t_{An}}), \mathord{\nf} B_{1}(\vec{t_{B1}}),\ldots,\mathord{\nf} B_{m}(\vec{t_{Bm}}),$ \\$\mathord{\nf} NH(\vec{t_H})$ if $H(\vec{t_H})$ is a DL-atom; or
\item[(2b.ii)] $H^{d}(\vec{t_H})\leftarrow A_{1}^{d}(\vec{t_{A1}}),\ldots, A_{n}^{d}(\vec{t_{An}}), \mathord{\nf} B_{1}(\vec{t_{B1}}),\ldots,\mathord{\nf} B_{m}(\vec{t_{Bm}})$\\ if $H(\vec{t_H})$ is a non-DL-atom.
\end{itemize}
\end{enumerate}
We define the \emph{doubled Hybrid MKNF knowledge base} $\cK^d=(\cO,\cO^d,\cP^d)$.
\end{definition}
Intuitively, we use an atom based on the original predicate $A$ to represent truth of $A$ in the original knowledge base, while the atom based on a newly introduced predicate $A^d$ represents non-falsity of $A$ in the original knowledge base, i.e., if we want to know whether some atom is (non-monotonically) false, then we query using the auxiliary predicate.
The new atom $NH(\vec{t_H})$ appearing in (2b.i), is used as a marker to distinguish between rules that may be affected by the derivability of the classical negation of its head (as in $\Gamma'_{\mathcal{K}}$) and the others (as in $\Gamma_{\mathcal{K}}$).
Note that this process of doubling the knowledge base has no impact on the (at best) polynomial data complexity of computing the well-founded MKNF model since it only alters the computation by the constant factor $2$.

\begin{example}
Consider $\cK$ from Example \ref{exPrel1}.
We obtain $\cK^d$ as follows.
\begin{align*}
\tt{C} & \sqsubseteq \tt{D} & \tt{C^d} & \sqsubseteq \tt{D^d}\\
\tt{C}\sqcap \tt{E} & \sqsubseteq \bot & \tt{C^d}\sqcap \tt{E^d} & \sqsubseteq \bot\\
\tt{C}(\tt{b}) & & \tt{C^d}(\tt{b}) &\\
\tt{p}(\tt{x}) & \leftarrow \nf \tt{D^d}(\tt{x}), \tt{o}(\tt{x}) & \tt{p^d}(\tt{x}) & \leftarrow \nf \tt{D}(\tt{x}), \tt{o^d}(\tt{x})\\
\tt{E}(\tt{x}) & \leftarrow \nf \tt{E^d}(\tt{x}), \tt{o}(\tt{x}) & \tt{E^d}(\tt{x}) & \leftarrow \nf \tt{E}(\tt{x}), \tt{o^d}(\tt{x}),\nf \tt{NE}(x)\\
\tt{o}(\tt{a}) &\leftarrow & \tt{o^d}(\tt{a}) & \leftarrow\\
\tt{o}(\tt{b}) &\leftarrow & \tt{o^d}(\tt{b}) & \leftarrow
\end{align*}
Only the rule with head $\tt{E^d}(\tt{x})$ contains the marker in the body as the original rule is the only one in $\cK$ with a DL-atom in the head.
Note that the atoms based on the predicate $\tt{o}$, which ensure DL-safety, could be excluded from the doubling for efficiency reasons.
\end{example}

The marker has to be referenced in the modified transform but, before that, we define a slightly different operator for doubled, positive Hybrid MKNF knowledge bases that takes into account the parallel computations on the two renamings of the ontology.

\begin{definition}\label{d:opRkDkTkd}
Let $\mathcal{K}^d=(\mathcal{O},\cO^d,\mathcal{P}^d)$ be a doubled, positive, ground Hybrid MKNF knowledge base.
The operators $R_{\mathcal{K}^d}$, $D_{\mathcal{K}^d}$, and $T_{\mathcal{K}^d}$ are defined on subsets of ${\sf KA}(\mathcal{K}^d)$ as follows:
\begin{align*}
R_{\mathcal{K}^d}(S) = & \,\{H \mid  \mathcal{P}^d \text{ contains a rule of the form $H\leftarrow A_1,\ldots A_n$} \\ 
& \,\text{ such that, for all }i, 1\leq i\leq n, A_{i}\in S\}\\
D_{\mathcal{K}^d}(S) = & \,\{\xi \mid \xi\in {\sf KA}(\mathcal{K})\text{ and }{\sf OB}_{\mathcal{O},S}\models \xi\} \,\cup \\
& \,\{\xi \mid \xi\in {\sf KA}(\mathcal{K}^d)\setminus {\sf KA}(\mathcal{K}) \text{ and }{\sf OB}_{\mathcal{O}^d,S}\models \xi\}\\
T_{\mathcal{K}^d}(S) = & \, R_{\mathcal{K}^d}(S)\cup D_{\mathcal{K}^d}(S)
\end{align*}
\end{definition}
These operator definitions are the same as those of
Definition~\ref{d:opRkDkTk} apart from two differences.  First, the
doubled knowledge base $\cK^d$ is considered and, consequently, atoms
from ${\sf KA}(\mathcal{K}^d)$ appear.  Second, the operator
$D_{\mathcal{K}^d}$ computes consequences from $\cO$ and $\cO^d$ in
parallel but limited to the corresponding set of atoms appearing in
each of the two renamings, thus preventing an inconsistency, e.g., in
$\cO$, from affecting the consistency of $\cO^d$.

Next, we present a slightly altered version of the MKNF-coherent transform (cf.\ Definition~\ref{d:MKNF-cohtransformK}) taking into account the doubled Hybrid MKNF knowledge base and the new negative literals of the form $NH(\vec{t_H})$ that serve as markers.

\begin{definition}\label{d:MKNF-cohtransformKd}
Let $\mathcal{K}^d=(\mathcal{O},\cO^d,\mathcal{P}^d)$ be a doubled, ground Hybrid MKNF knowledge base and $S\subseteq{\sf KA}(\mathcal{K}^d)$.
The \emph{MKNF$^d$-coherent transform} $\mathcal{K}^d//' S$ is defined as $\mathcal{K}^d//' S=(\mathcal{O},\cO^d,\mathcal{P}^d//' S)$, where $\mathcal{P}^d//' S$ contains all rules $H\leftarrow A_{1},\ldots, A_{n}$ for which there exists a rule 
\[
H\leftarrow A_{1}, \ldots, A_{n}, \nf B_{1},\ldots, \nf B_{m}
\]
in $\mathcal{P}^d$ with 
\begin{enumerate}
\item $B_{j}\not\in S$ for all $1\leq j\leq m$; and 
\item${\sf OB}_{\mathcal{O},S}\not\models \neg H_1(\vec{t_{H_1}})$ if $\nf NH_1(\vec{t_{H_1}})$ appears in the body, where $H=H_1^d(\vec{t_{H_1}})$.
\end{enumerate}
\end{definition}
This definition is almost identical to the transformation of Definition~\ref{d:MKNF-cohtransformK} with the only difference that the removal due to classical negation is only possible in marked rules.
Given Definition~\ref{d:doubling}, this means that only rules whose head is a DL-atom and built by means of a doubled predicate may be eliminated that way (case 2b.i of Definition~\ref{d:doubling}).
Additionally, as we will see in Section \ref{s:SLGO}, the marker itself can be used to actually trigger a query to the ontology for the classical negation of the atom in the head (using the original predicate for the atom).

We can now define a new operator $\Gamma_{\cK}^{d}$ for ground knowledge bases $\cK$ similar to the ones in Definition \ref{d:GammaK}, but that operates on atoms of ${\sf KA}(\mathcal{K}^{d})$. 

\begin{definition}\label{d:GammaKd}
Let $\mathcal{K}^d=(\mathcal{O},\cO^d,\mathcal{P}^d)$ be a doubled, ground Hybrid MKNF knowledge base and $S\subseteq {\sf KA}(\mathcal{K}^{d})$.
We define:
\[
\Gamma_{\mathcal{K}^d}(S) =  \lfp(T_{\mathcal{K}^{d}//'S}).
\]
\end{definition}

We can show that this operator is antitonic just as its two predecessors.

\begin{lemma}\label{l:GammaKdanti}
Let $\mathcal{K}^d$ be a doubled, ground Hybrid MKNF knowledge base and $S_1\subseteq S_2\subseteq {\sf KA}(\mathcal{K}^{d})$.
Then $\Gamma_{\mathcal{K}^d}(S_2) \subseteq \Gamma_{\mathcal{K}^d}(S_1)$.
\end{lemma}
\begin{proof}
We have to show that $\lfp(T_{\mathcal{K}^{d} //'S_2})\subseteq \lfp(T_{\mathcal{K}^{d} //'S_1})$.
\noindent
Since $\mathcal{K}$ is finite, $\mathcal{K}^{d}$ is also finite, and we prove by induction on $n$ that $T_{\mathcal{K}^{d} //'S_2}\uparrow n\subseteq T_{\mathcal{K}^{d} //'S_1}\uparrow n$ holds.

The base case for $n=0$ is trivial since $\emptyset\subseteq\emptyset$.

Assume that $T_{\mathcal{K}^{d}//'S_2}\uparrow n \subseteq T_{\mathcal{K}^{d}//'S_1}\uparrow n$ holds, consider $H \in T_{\mathcal{K}^{d} //'S_2}\uparrow (n+1)$.
Then $H \in T_{\mathcal{K}^{d}//'S_2}(T_{\mathcal{K}^{d} //'S_2}\uparrow n)$ and there are two cases to consider: 
\begin{enumerate}
\item $\mathcal{K}^{d} //'S_2$ contains a rule of the form $H \leftarrow A_{1},\ldots , A_{n}$ such that $A_{i}\in T_{\mathcal{K}^{d} //'S_2}\uparrow n$ for each $1\leq i \leq n$.
In this case, since $S_1\subseteq S_2$ holds, we also have $H \leftarrow A_{1},\ldots ,A_{n}$ in $\mathcal{K}^{d} //'S_1$ and, by the induction hypothesis, $A_{i}\in T_{\mathcal{K}^{d} //'S_1}\uparrow n$ holds for each $1\leq i \leq n$.
Hence, $H \in T_{\mathcal{K}^{d} //'S_1}\uparrow (n+1)$.
\item $H$ is a consequence obtained from $D_{\mathcal{K}^{d}}$.
But $D_{\mathcal{K}^{d}}$ derives only consequences from the unchanged DL renamings $\mathcal{O}$ and $\cO^d$ together with $T_{\mathcal{K}^{d} //'S_2}\uparrow n$. By the induction hypothesis, we conclude that $H \in T_{\mathcal{K}^{d} //'S_1}\uparrow (n+1)$.
\end{enumerate}
This finishes the proof.
\end{proof}

Since this new operator is antitonic, we can define its iteration similar to Definition \ref{d:ite}, but now with just one operator.

\begin{definition}\label{d:itealt}
Let $\mathcal{K}^d$ be a doubled, ground Hybrid MKNF knowledge base.
We define:
\begin{align*}
\mathbf{P}^{d}_{0} & = \emptyset & \mathbf{N}^{d}_{0} & = {\sf KA}(\mathcal{K}^{d})\\
\mathbf{P}^{d}_{n+1} & = \Gamma_{\mathcal{K}^d}(\mathbf{N}^{d}_{n}) &  \mathbf{N}^{d}_{n+1} & = \Gamma_{\mathcal{K}^d}(\mathbf{P}^{d}_{n}) \\
\mathbf{P}^{d}_{\omega} & = \bigcup\mathbf{P}^{d}_{n} &\mathbf{N}^{d}_{\omega} & =  \bigcap\mathbf{N}^{d}_{n}
\end{align*}
\end{definition}

The correspondence between Definitions \ref{d:ite} and \ref{d:itealt} can be established with a precise relation between the atoms in the doubled knowledge base $\mathcal{K}^{d}$ and those in $\mathcal{K}$.
To ease the proof of the corresponding property, we rely on an adaptation of the notion of unfounded sets to doubled Hybrid MKNF KBs.
For that purpose, we also adapt the notion of dependency.

\begin{definition}\label{d:depOBd}
Let $\mathcal{K}^d=(\cO,\cO^d,\mathcal{P}^d)$ be a doubled, ground Hybrid MKNF knowledge base, $H$ an atom with $H\in {\sf KA}(\mathcal{K}^d)$, and $S$ a (possibly empty) set of atoms with $S\subseteq {\sf KA}(\mathcal{K}^d)$.
We say that $H$ \emph{depends} on $S$ if and only if, for $\cO'=\cO$ or $\cO'=\cO^d$:
\begin{itemize}
\item[(i)] ${\sf OB}_{\mathcal{O}',S}\models H$ and
\item[(ii)] there is no $S'$ with $S'\subset S$ such that ${\sf OB}_{\mathcal{O}',S'}\models H$.
\end{itemize}
\end{definition}

Based on this notion of dependency, the notion of an unfounded set for Hybrid MKNF is extended from Definition~\ref{d:unfMKNF} to include $\cO$ and $\cO^d$ of the doubled KB.

\begin{definition}\label{d:unfMKNFd}
Let $\cK^d=(\mathcal{O},\cO^d,\mathcal{P}^d)$ be a doubled, ground Hybrid MKNF knowledge base and $(T,F)$ a pair of sets such that $T,F\subseteq {\sf KA}(\mathcal{K}^d)$.
We say that $U\subseteq {\sf KA}(\mathcal{K}^d)$ is an \emph{unfounded set} (of $\cK^d$) \emph{with respect to $(T,F)$} if, for each atom $H\in U$, the following conditions are satisfied: 
\begin{itemize}
\item[(Ui)] for each rule $H \leftarrow \mathcal{B}$ in $\mathcal{P}$ at least one of the following holds.
\begin{itemize}
\item[(Uia)] Some atom $A$ appears in $\mathcal{B}$ and in $U\cup F$. 
\item[(Uib)] Some negative literal $\nf B$ appears in $\mathcal{B}$ and in $T$.
\item[(Uic)] ${\sf OB}_{\mathcal{O},T}\models \neg H_1(\vec{t_{H_1}})$ and $\nf NH_1(\vec{t_{H_1}})\in \mathcal{B}$, where $H=H_1^d(\vec{t_{H_1}})$
\end{itemize} 
\item[(Uii)] for each (possibly empty) $S$ on which $H$ depends, with $S\subseteq {\sf KA}(\mathcal{K})$ and ${\sf OB}_{\mathcal{O},S}$ consistent, there is at least one atom $A$ such that ${\sf OB}_{\mathcal{O},S\setminus \{ A\}}\not\models H$ and $A$ in $U\cup F$.
\item[(Uii$^d$)] for each (possibly empty) $S$ on which $H$ depends, with $S\subseteq {\sf KA}(\mathcal{K}^d)$ and ${\sf OB}_{\mathcal{O}^d,S}$ consistent, there is at least one atom $A$ such that ${\sf OB}_{\mathcal{O}^d,S\setminus \{ A\}}\not\models H$ and $A$ in $U\cup F$.
\end{itemize}
The union of all unfounded sets of $\cK^d$ w.r.t.\ $(T,F)$ is called the \emph{greatest unfounded set} of $\cK^d$ w.r.t.\ $(T,F)$ and denoted $U_{\mathcal{K}^d}(T,F)$.
\end{definition} 
Of course, (Uii$^d$) is just a copy of (Uii) to deal with $\cO^d$, the copy of $\cO$. 

The correspondence to the sequence $\mathbf{N}^d_i$ for all $i$ can now be established.

\begin{lemma}\label{l:convAltWd}
Let $\mathcal{K}^d=(\mathcal{O},\mathcal{O}^d,\mathcal{P}^d)$ be a
doubled, ground Hybrid MKNF knowledge base and
$(\mathbf{P}^d_i,\mathbf{N}^d_i)$ a pair of sets such that $\mathbf{P}^d_i,\mathbf{N}^d_i \in
{\sf KA}(\mathcal{K}^d)$ in the computation of the alternating
fixpoint of $\mathcal{K}^d$  (Definition~\ref{d:itealt}).  Then the following holds:
\[
{\sf KA}(\mathcal{K}^d)\setminus\mathbf{N}^d_{i+1} = U_{\mathcal{K}^d}(\mathbf{P}^d_i,{\sf KA}(\mathcal{K}^d)\setminus\mathbf{N}^d_i)
\]
\end{lemma}
\begin{proof}
We show both inclusions from which the equality follows.

${\sf KA}(\mathcal{K}^d)\setminus\mathbf{N}^d_{i+1} \subseteq U_{\mathcal{K}^d}(\mathbf{P}^d_i,{\sf KA}(\mathcal{K}^d)\setminus\mathbf{N}^d_i)$:
Let $H\in {\sf KA}(\mathcal{K}^d)\setminus\mathbf{N}^d_{i+1}$.
Then $H\not\in \mathbf{N}^d_{i+1}$, i.e., $H\not\in \Gamma_{\mathcal{K}^d}(\mathbf{P}^d_i)$ and $H\not\in \lfp(T_{\mathcal{K}^d//'\mathbf{P}^d_i})$.
Thus, two conditions hold.
First, for all rules of the form $H\leftarrow \mathcal{B}^+ \wedge \mathcal{B}^-$ in $\mathcal{P}^d$, there is at least one $A\in \mathcal{B}^+$ with $A\in {\sf KA}(\mathcal{K}^d)\setminus \mathbf{N}^d_{i+1}$, or at least one $\nf B\in \mathcal{B}^-$ with $B \in \mathbf{P}^d_i$, or ${\sf OB}_{\mathcal{O},\mathbf{P}_i}\models \neg H_1(\vec{t_{H_1}})$ and $\nf H_1(\vec{t_{H_1}})\in \mathcal{B}^-$, where $H=H_1^d(\vec{t_{H_1}})$.
Second, neither ${\sf OB}_{\mathcal{O},\mathbf{N}^d_{i+1}}\models H$ nor ${\sf OB}_{\mathcal{O}^d,\mathbf{N}^d_{i+1}}\models H$ holds.
The first condition corresponds exactly to (Ui) of Definition~\ref{d:unfMKNFd} w.r.t.\ $(\mathbf{P}^d_i,{\sf KA}(\mathcal{K}^d)\setminus \mathbf{N}^d_i)$.
We derive from the second condition that, for all $S$ with $S\subseteq {\sf KA}(\mathcal{K}^d)$ on which $H$ depends, there is at least one atom $A$ such that ${\sf OB}_{\mathcal{O},S\setminus \{A\}}\not\models H$, respectively ${\sf OB}_{\mathcal{O}^d,S\setminus \{A\}}\not\models H$, and $A$ in ${\sf KA}(\mathcal{K}^d)\setminus \mathbf{N}^d_{i+1}$.
This matches condition (Uii) (resp.\ condition (Uii$^d$) of Definition~\ref{d:unfMKNFd} w.r.t.\ $(\mathbf{P}^d_i,{\sf KA}(\mathcal{K}^d)\setminus \mathbf{N}^d_i)$, and we conclude that $H\in U_{\mathcal{K}^d}(\mathbf{P}^d_i,{\sf KA}(\mathcal{K}^d)\setminus \mathbf{N}^d_i)$.

$U_{\mathcal{K}^d}(\mathbf{P}^d_i,{\sf KA}(\mathcal{K}^d)\setminus\mathbf{N}^d_i)\subseteq {\sf KA}(\mathcal{K}^d)\setminus\mathbf{N}^d_{i+1}$:
Let $H\in U_{\mathcal{K}^d}(\mathbf{P}^d_i,{\sf KA}(\mathcal{K}^d)\setminus\mathbf{N}^d_i)$.
Then $H$ occurs in the greatest unfounded set w.r.t.\ $(\mathbf{P}^d_i,{\sf KA}(\mathcal{K}^d)\setminus\mathbf{N}^d_i)$.
It follows from Definition~\ref{d:unfMKNFd} that $H\not\in\mathbf{N}^d_{i+1}$.
Consequently, $H\in {\sf KA}(\mathcal{K}^d)\setminus\mathbf{N}^d_{i+1}$.
\end{proof}

We can now show the correspondence between atoms of $\cK$ in
the fixed point of Definition~\ref{d:ite} and those of $\cK^d$ in
the fixed point of Definition~\ref{d:itealt}.

\begin{proposition}\label{p:altcomp}
Let $\mathcal{K}=(\mathcal{O},\mathcal{P})$ be a ground Hybrid MKNF knowledge base.
Then the following holds:
\begin{itemize}
\item $A\in \mathbf{P}_{\omega}$ if and only if $A \in \mathbf{P}^{d}_{\omega}$.
\item $B \not\in \mathbf{N}_{\omega}$ if and only if $B^{d} \not\in \mathbf{N}^{d}_{\omega}$. 
\end{itemize} 
\end{proposition}

\begin{proof}
We show by induction on $n$ that two conditions hold.
\begin{itemize}
\item[(i)] $A\in \mathbf{P}_{n}$ if and only if $A \in \mathbf{P}^{d}_{n}$
\item[(ii)] $B \not\in \mathbf{N}_{n}$ if and only if $B^{d} \not\in \mathbf{N}^{d}_{n}$ 
\end{itemize} 
This is sufficient since the grounded knowledge base is finite, which means that the iteration is finite and stops for some natural number $n$, i.e., the two fixpoints coincide on the relevant atoms as in (i) and (ii).

The base case for $n=0$ is straightforward since $\mathbf{P}_{0}$ and $\mathbf{P}^{d}_{0}$ are empty while $\mathbf{N}_{0}$ and $\mathbf{N}^{d}_{0}$ both contain their entire Herbrand base.

(1) So, suppose that (i) and (ii) hold for $n$ and let $A\in  \mathbf{P}_{n+1}$ and $B \not\in \mathbf{N}_{n+1}$.
We show that $A\in  \mathbf{P}^{d}_{n+1}$ and $B^{d} \not\in \mathbf{N}^{d}_{n+1}$.
The other direction of the equivalence follows from an identical argument.

\begin{itemize}
\item[(i)] First, suppose that $A \in \mathbf{P}_{n+1}$ but $A\not\in \mathbf{P}_{n}$ (otherwise we obtain the result by the induction hypothesis (1) immediately).
We show that $A \in \mathbf{P}^{d}_{n+1}$. 
If $A \in \mathbf{P}_{n+1}$, then $A\in\Gamma_{\mathcal{K}}(\mathbf{N}_{n})$, by Definition~\ref{d:ite}, and, thus, $A\in \lfp(T_{\mathcal{K}/\mathbf{N}_{n}})$ by Definition~\ref{d:GammaK}.
So $A\in T_{\mathcal{K}/\mathbf{N}_{n}}\uparrow m$ for some $m$ and we show by induction on $m$ that  $A\in T_{\mathcal{K}^{d}//'\mathbf{N}_{n}}\uparrow m$ (2), which implies that $A\in \mathbf{P}^{d}_{n+1}$.
The base case for $m=0$ holds immediately.
Assume the claim (2) holds for $m$, we show it for $m+1$.
Suppose that $A\in T_{\mathcal{K}/\mathbf{N}_{n}}\uparrow (m+1)$ then $A\in T_{\mathcal{K}/\mathbf{N}_{n}}(T_{\mathcal{K}/\mathbf{N}_{n}}\uparrow m)$.
Then either $A\in R_{\mathcal{K}/\mathbf{N}_{n}}(T_{\mathcal{K}/\mathbf{N}_{n}}\uparrow m)$ or $A\in D_{\mathcal{K}/\mathbf{N}_{n}}(T_{\mathcal{K}/\mathbf{N}_{n}}\uparrow m)$.
We start with the first case, i.e., there is a rule $A\leftarrow A_{1},\ldots,A_{n}, \nf B_{1}, \ldots, \nf B_{m}$ with $A_{i}\in T_{\mathcal{K}/\mathbf{N}_{n}}\uparrow m$ and $\nf B_{j}\not\in \mathbf{N}_{n}$ for all $i$ and $j$.
For each such rule $A\leftarrow A_{1},\ldots,A_{n}, \nf B_{1}, \ldots, \nf B_{m}$ in $\mathcal{K}$ there is, according to Definition \ref{d:doubling}, a rule $A\leftarrow A_{1},\ldots, A_{n}, \nf B^{d}_{1}, \ldots, \nf B^{d}_{m}$ in $\mathcal{P}^{d}$.
Since, by the induction hypothesis (1), we have that $B_{i} \not\in \mathbf{N}_{n}$ if and only if $B_{i}^{d} \not\in \mathbf{N}^{d}_{n}$ we obtain that each rule in  $\mathcal{K}/\mathbf{N}_{n}$ has its correspondent in $\mathcal{K}^{d}//'\mathbf{N}^{d}_{n}$.
We obtain by the nested induction hypothesis of (2) that  $A\in T_{\mathcal{K}^{d}//'\mathbf{N}_{n}}\uparrow (m+1)$.
Otherwise, $A\in D_{\mathcal{K}/\mathbf{N}_{n}}(T_{\mathcal{K}/\mathbf{N}_{n}}\uparrow m)$ holds, and $A\in D_{\mathcal{K}^{d}//'\mathbf{N}_{n}}(T_{\mathcal{K}^{d}//'\mathbf{N}_{n}}\uparrow m)$ is obtained immediately by the induction hypothesis (2) and the identical ontologies $\mathcal{O}$ contained in $\cK^d$ and $\cK$.

\item[(ii)] To prove (ii) we suppose as well that $B \not\in \mathbf{N}_{n+1}$ but $B \in \mathbf{N}_{n}$.
We show that $B^{d} \not\in \mathbf{N}^d_{n+1}$.
If $B \not\in \mathbf{N}_{n+1}$, then $B \in U_{\mathcal{K}}(\mathbf{P}_n,{\sf KA}(\mathcal{K})\setminus\mathbf{N}_n)$.
By Definitions~\ref{d:unfMKNFd}, and \ref{d:doubling}, we obtain that $B \in U_{\mathcal{K}^d}(\mathbf{P}^d_n,{\sf KA}(\mathcal{K}^d)\setminus\mathbf{N}^d_n)$.
Hence, by Lemma~\ref{l:convAltWd}, $B\not\in \mathbf{N}^d_{n+1}$.
\end{itemize}
This finishes the proof.
\end{proof}

It follows immediately from this proposition that we can use this alternative computation to compute the well-founded MKNF model.
Formally we obtain the following theorem, which shows the adapted well-founded MKNF model.

\begin{theorem}\label{t:nMKNFWFS}
Let $\mathcal{K}=(\mathcal{O},\mathcal{P})$ be a ground,
MKNF-consistent Hybrid MKNF knowledge base and let
$\mathbf{P}^{d}_{\mathcal{K}}, \mathbf{N}^{d}_{\mathcal{K}} \subseteq
{\sf KA}(\mathcal{K}^{d})$ with
\begin{tabbing}
$\mathbf{P}^{d}_{\mathcal{K}}=\{A\mid A\in \mathbf{P}^{d}_{\omega}$ and  $A\in {\sf KA}(\mathcal{K})\}$,$\mathbf{N}^{d}_{\mathcal{K}}= \{A^{d} \mid
A^{d}\in \mathbf{N}^d_{\omega}$ and $A^d\in {\sf KA}(\mathcal{K}^d)\}$.
\end{tabbing}
Then
\[
M_{WF} = \{A\mid A\in \mathbf{P}^{d}_{\mathcal{K}}\}\cup \{\pi(\mathcal{O})\}\cup \{\nf A \mid A^{d}\in ({\sf KA}(\mathcal{K}^{d})\setminus {\sf KA}(\mathcal{K})) \setminus\mathbf{N}^{d}_{\mathcal{K}}\}
\]
is the well-founded MKNF model of $\mathcal{K}$.
\end{theorem}
\begin{proof}
The result is an immediate consequence of Proposition~\ref{p:altcomp} and Definition \ref{d:MKNFWFS}.
\end{proof}
The two sets $\mathbf{P}^{d}_{\mathcal{K}}$ and $\mathbf{N}^{d}_{\mathcal{K}}$ are just used to remove superfluous atoms, e.g., the atoms based on doubled predicates for $\mathbf{P}^{d}_{\mathcal{K}}$.
We note that for practical purposes we also derive from this theorem
and Proposition~\ref{p:altcomp} that we have to use the new predicates
$A^{d}$ if we query for negative literals.

To better illustrate how each  of the two computations work, we finish the section with a technical example.
\begin{example}\label{e:double}
Consider the knowledge base $\cK$. 
\begin{align*}
\tt{Q} & \sqsubseteq  \neg \tt{R} \\
\tt{p}(\tt{a}) & \leftarrow  \nf \tt{p}(\tt{a}) \\
\tt{Q}(\tt{a}) & \leftarrow  \\
\tt{R}(\tt{a}) & \leftarrow \nf \tt{R}(\tt{a})
\end{align*}
We now show how both computation work in this example, yielding (as expected from Proposition \ref{p:altcomp}) the same results.

We can compute the two sequences $\mathbf{P}_{i}$ and $\mathbf{N}_{i}$ and obtain:
\begin{align*}
\mathbf{P}_{0}&=\emptyset & \mathbf{N}_{0}&=\{\tt{p}(\tt{a}),\tt{Q}(\tt{a}),\tt{R}(\tt{a})\}\\
\mathbf{P}_{1}&= \{\tt{Q}(\tt{a})\} & \mathbf{N}_{1} &= \mathbf{N}_{0}\\
\mathbf{P}_{2}&=\mathbf{P}_{1} & \mathbf{N}_{2} &= \{\tt{p}(\tt{a}),\tt{Q}(\tt{a})\}\\
\mathbf{P}_{3}&=\{\tt{p}(\tt{a}),\tt{Q}(\tt{a}),\tt{R}(\tt{a})\} & \mathbf{N}_{3} &= \mathbf{N}_{2}\\
\mathbf{P}_{4}&=\mathbf{P}_{3}& \mathbf{N}_{4} &= \emptyset
\end{align*}
The knowledge base is obviously MKNF-inconsistent since we derive that everything is true and false at the same time.

Now we apply the alternative computation using the doubled set of rules $\cP^d$ and the ontology $\cO$ and its renaming $\cO^d$ including the special marker predicates $\tt{NR}$ and $\tt{NQ}$.
\begin{align*}
\tt{Q}  & \sqsubseteq  \neg \tt{R} & \tt{Q}^{\tt{d}}  & \sqsubseteq  \neg \tt{R}^{\tt{d}} \\
\tt{p}(\tt{a}) & \leftarrow  \nf \tt{p}^{\tt{d}}(\tt{a}) & \tt{p}^{\tt{d}}(\tt{a})  & \leftarrow  \nf \tt{p}(\tt{a})\\
\tt{Q}(\tt{a}) & \leftarrow  & \tt{Q}^{\tt{d}}(\tt{a})  & \leftarrow  \nf \tt{NQ}(\tt{a})\\
\tt{R}(\tt{a}) & \leftarrow  \nf \tt{R}^{\tt{d}}(\tt{a}) & \tt{R}^{\tt{d}}(\tt{a})  & \leftarrow  \nf \tt{R}(\tt{a}), \nf \tt{NR}(\tt{a})
\end{align*}
We compute the two sequences for the transformed knowledge base $\cK^d$ and obtain:
\begin{align*}
\mathbf{P}^{d}_{0}& =\emptyset & \mathbf{N}^{d}_{0} &=\{\tt{p}(\tt{a}),\tt{p}^{\tt{d}}(\tt{a}),\tt{Q}(\tt{a}),\tt{Q}^{\tt{d}}(\tt{a}),\tt{R}(\tt{a}), \tt{R}^{\tt{d}}(\tt{a}),\tt{NQ}(\tt{a}),\tt{NR}(\tt{a})\}\\
\mathbf{P}^{d}_{1}& = \{\tt{Q}(\tt{a}),\tt{Q}^{\tt{d}}(\tt{a})\} & \mathbf{N}^{d}_{1} &= \mathbf{N}_{0}^d\\
\mathbf{P}^{d}_{2}& =\mathbf{P}^d_{1} & \mathbf{N}^{d}_{2} &= \{\tt{p}(\tt{a}), \tt{p}^{\tt{d}}(\tt{a}), \tt{Q}(\tt{a}), \tt{Q}^{\tt{d}}(\tt{a}), \tt{R}(\tt{a})\} \\
\mathbf{P}^{d}_{3}& =\{\tt{p}(\tt{a}),\tt{Q}(\tt{a}),\tt{R}(\tt{a})\} & \mathbf{N}^{d}_{3} &= \mathbf{N}^d_{2}\\
\mathbf{P}^{d}_{4}& =\mathbf{P}^d_{3}& \mathbf{N}^{d}_{4} &= \{\tt{p}(\tt{a}),\tt{Q}(\tt{a}),\tt{R}(\tt{a})\} 
\end{align*}
All atoms based on original predicates are true while all doubled atoms are false.
This indicates again that the knowledge base is MKNF-inconsistent.
\end{example}

Note that the inconsistency in $\tt{R}$ ensures that everything in the
knowledge base is considered inconsistent.  This does not always hold
(consider adding a fact $\tt{p}(\tt{a})\leftarrow$ to the rules, then
$\tt{p}^{\tt{d}}(\tt{a})$ is not false but true).  However, whenever
we encounter an atom such that $\tt{P}$ is true while
$\tt{P}^{\tt{d}}$ is false, then we know that the KB is
MKNF-inconsistent.  Adapting Theorem \ref{t:consKB} to this
alternative computation of the well-founded MKNF model is not trivial,
since the computation is now more intertwined.  But this does not
constitute a problem.  The purpose of this computation is to provide a
link to top-down querying, where, for reasons of efficiency, we do not
want to test whether the entire KB is MKNF-consistent:
we only consider the portion of the KB used in the derivation of the considered query.

\section{Tabled SLG($\cO$)-resolution for Hybrid MKNF}\label{s:SLGO}

We present \SLGO{} for Hybrid MKNF knowledge bases which extends SLG
resolution from \cite{CheW96} with an oracle to capture first-order
deduction in DLs.  SLG evaluation models well-founded computation for
logic programs at an operational level, ensuring goal-directedness,
termination and optimal complexity for a large class of programs
(cf.\ \cite{CheW96}). 
At the same time it has motivated the design of modern tabling
engines, and captures many aspects of their behavior.  When SLG is
extended with an oracle in \SLGO{}, several of the definitions of SLG
are affected.  In this section we present the definitions of \SLGO{},
as well as defining when an oracle is suitable for use in an
evaluation.  As the \SLGO{} definitions are presented, we make clear
how they differ from those of SLG. 
For the definition of \SLGO{}, we follow and extend the model of \cite{Swif99b}.

Briefly, an \SLGO{} evaluation is a sequence of forests (sets) of
program trees.
Program trees
themselves correspond to subgoals that have been encountered in an
evaluation.  The nodes in these trees contain sets of literals divided
into those literals that have not been examined, and others that have
been examined, but their resolution delayed (cf.\
Definition~\ref{def:forest}).  
The need to delay some literals arises
for the following reason.  
Modern Prolog engines rely on a fixed order
for selecting literals in a rule, e.g., always left-to-right.
However, well-founded computations cannot be performed using a
fixed-order literal selection function.\footnote{A literal selection
function is employed to choose the next literal to resolve in the body
of a rule. In \SLGO{}, the only requirement for a selection function
is that DL-atoms are not selected until they are ground, which is
always possible given DL-safety of conjunctive queries and the rules
appearing in the knowledge base (cf. Definition \ref{d:dlsafecq}).}
Hence, in \SLGO{} the {\sc delay} operation may postpone evaluation of
some literals, which may be later resolved through an operation called
{\sc simplification}.  In addition to modeling the operational
behavior of Prolog, the use of delay and simplification supports the
termination and complexity results of \SLGO{} discussed in
Section~\ref{s:prop}, analogous to those presented for SLG in \cite{CheW96}.

\begin{example}\label{e:wf2}
To ease the understanding of \SLGO{}, we present a
concrete example of an \SLGO{} evaluation that does not use an oracle.
Consider the following Hybrid MKNF knowledge base $\mathcal{K}$ with empty $\cO$.
\begin{align}
\tt{p(b)} & \leftarrow \label{r1}\\
\tt{p(c)} & \leftarrow \nf \tt{p(a)}\label{r2} \\
\tt{p(X)} & \leftarrow \tt{t(X,Y,Z)},\nf \tt{p(Y)}, \nf \tt{p(Z)}\label{r3} \\
\tt{p(a)} & \leftarrow \tt{p(b)},\tt{p(a)}\label{r4} \\
\tt{t(a,a,b)} & \leftarrow \label{r5}\\
\tt{t(a,b,a)} & \leftarrow \label{r6}
\end{align}
We consider the
query {\tt p(c)} to $\cK$ in which none of the atoms is a DL-atom, i.e., no oracle needs to be used.
The \SLGO{} forest at the end of this evaluation is shown in
Figure~\ref{fig:plain-forest} where each node is labeled with a number
indicating the order in which it was created in the \SLGO{}
evaluation.  Nodes consist of either the symbol \emph{fail}, or of
a head representing the bindings
made to an atomic subgoal and a body with a set of {\em
  Delays}, followed by the $|$ symbol, followed by {\em Goals} that are still to be examined.  The
evaluation begins by creating a tree for the initial query with root
{\tt p(c)}$\leftarrow$ {\tt $|$p(c)} in node 1. Children of root
nodes are created via the operation \pgmcr{} just as in the SLD
resolution of Prolog.  Accordingly, the evaluation uses rule
(\ref{r2}) to create node 2. 
The (only possible) literal $\nf$ {\tt p(a)} in node 2 is selected. 
This literal has an underlying subgoal {\tt p(a)} that does not correspond to the
root of any tree in the forest so far.  Thus, the \SLGO{} operation
\newsg{} creates a new tree for {\tt p(a)} (node 3), whose child, node
4, is created by \pgmcr{} using rule (\ref{r3}).  The \newsg{}
operation is again used to create a new tree for the selected literal
{\tt t(a,X,Y)} (node 5), and children nodes 6 and 7 are created by \pgmcr{} from
rules (\ref{r5}) and (\ref{r6}). These latter nodes have empty {\em Goals}
and are termed {\em answers}; moreover, since they also have empty
{\em Delays}, they are {\em unconditional} answers.\footnote{In a practical
  program, a predicate defined by simple facts would not be evaluated
  using tabling, but rather would use SLD resolution as in Prolog.}
Any atom in the ground instantiation of an unconditional answer is true in the well-founded MKNF model, cf.\  Theorem~\ref{th:correct}. 
The \SLGO{} operation \ansret{} is used to
resolve the first of these answers against the selected literal of
node 4, producing node 9.  The selected literal of this latter node
has {\tt p(a)} as its underlying subgoal, but there is already a tree
for {\tt p(a)} in the forest and there are no answers for {\tt p(a)}
to return.  
Since there is another unconditional answer for {\tt t(a,X,Y)} (in node 7), \anscr{} can be used to produce node 10. 
The underlying subgoal {\tt p(b)} is selected, by  \newsg{} the 
tree for {\tt p(b)} is created, and it is eventually determined that
the subgoal {\tt p(b)} has an unconditional answer (node 12);
accordingly, using the \negfail{} operation, the {\em failure node}, node 14, is created.
Then, the computation, via \pgmcr{} and program rule (\ref{r4}), 
produces another child for {\tt p(a)}, node 15, and resolves
{\tt p(b)} (node 16).  At this stage the subgoal {\tt p(a)} is neither true,
as no unconditional answers have been derived for it, nor false as one
of its possible derivations, node 9, effectively has a loop through
negation.  However, in \SLGO{} it is possible to apply the {\sc delay} operation
to the selected negative literal, by moving it from the {\em Goals} to
the right of the $|$ symbol into the {\em Delays} to the left of the
$|$ symbol.  This {\sc delay} operation produces node 17, which is
termed a {\em conditional} answer, as it has empty {\em Goals} but
non-empty {\em Delays}.\footnote{Choosing delay in this order is not
optimal and is made for purposes of illustrating the operations of
\SLGO. 
This does not affect the result of the query itself since \SLGO{} is shown to be confluent 
in Theorem~\ref{t:appord}.}  {\sc Delay} also produces node 18 whose new selected
literal {\tt not p(b)} now fails (given the unconditional answer in node 12), 
producing the failure node 19.  
At this stage, all possible operations for non-answer nodes in {\tt p(a)}
and the trees it depends on have been performed so that {\tt p(a)} may
be {\em completed} (step 20).  The completed subgoal {\tt p(a)} has no
answers, and so is termed {\em failed} and is false in the
well-founded MKNF model of $\cK$.  This failed literal can be removed from
the delay list of node 18 through the {\sc simplification}
operation producing the unconditional answer node 21.

\begin{center}
\begin{figure}[t]
\includegraphics[width=13cm]{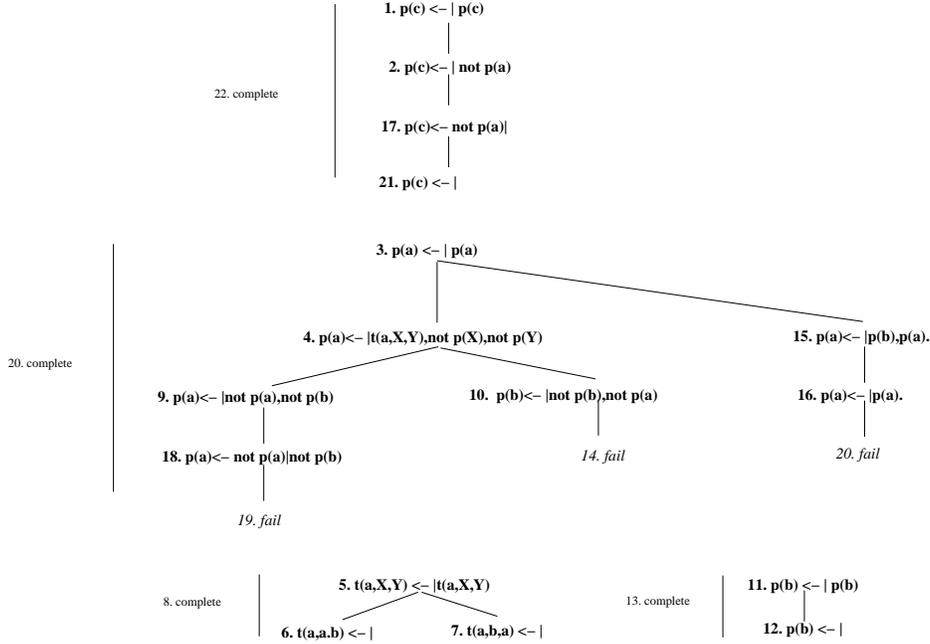}
\caption{Final forest for the query $\tt{p(c)}$ to $P_1$.}\label{fig:plain-forest}
\end{figure} 
\end{center}
\end{example}

Example~\ref{e:wf2} covers most of the main aspects of \SLGO{}, more precisely the main aspect of the
underlying formalism, SLG, that is applicable to normal logic programs.
SLG does not especially differ from other
Prolog-like tabling formalisms in the case of programs that do not use default negation
($\mathbf{not}$). However, for negation it introduces the
concept of delaying literals in order to be able to find witnesses of
failure anywhere in a rule, along with the concept of simplifying
these delayed literals whenever their truth value becomes known.
\SLGO{} allows additionally the usage of an oracle to incorporate reasoning
in the DL part $\cO$, and we present its definitions in the following.

An \SLGO{} evaluation proceeds by constructing a forest
according to the set of \SLGO{} operations.  Such a forest, and the
trees and nodes it contains are defined as follows:
\begin{definition}\label{def:forest}
A \emph{node} has the form
\[
AnswerTemplate \mif{} Delays | Goals \qquad \textrm{or} \qquad \emph{fail}.
\]
In the first form, $AnswerTemplate$ is an atom or a classically
negated atom, while $Delays$ and $Goals$ are sequences of literals.
The second form is called a {\em failure node}.
A \emph{program tree} $T$ is a tree of nodes whose root is of the form
$S \mif{} |S$ for some atom $S$ or a classically negated atom $S=\neg S_1$: we call $S$ {\em the root node for
$T$} and {\em $T$ the tree for $S$}.  An \emph{\SLGO{} forest}
$\mathcal{F}$ is a set of program trees.
A node $N$ is an {\em answer} when it is a leaf node for which $Goals$
is empty.  If the $Delays$ of an answer is empty, it is termed an {\em
unconditional answer}, otherwise, it is a {\em conditional answer}.  A
program tree $T$ may be marked with the symbol {\em complete}.
\end{definition}

The notions in Definition~\ref{def:forest} are almost identical to
previous formulations of SLG resolution. The only difference 
is that \SLGO{} allows for the appearance of classically
negated atoms as roots to incorporate possible calls for the classical
negation as required by the bottom-up computation in
Definition~\ref{d:MKNF-cohtransformKd}.  Such a literal $\neg A$ only
appears in an {\em AnswerTemplate} or as the only goal in the root
node, and is only used to query $\cO$.\\

An \SLGO{} evaluation of a query $Q$ starts with an initial forest with just one node $Q \mif{} | Q$ and creates a sequence of forests.
Each forest is obtained from the previous one by applying one \SLGO{} operation.
If no further \SLGO{} operation is applicable, then the final forest for the evaluation of the query has been reached. 
We introduce these \SLGO{} operations incrementally, in Definitions \ref{def:cass-ops-1}, \ref{def:cass-ops-2}, \ref{def:cass-ops-3}, and \ref{def:cass-ops-4}.  
But before we present the first set of operations, we need two auxiliary definitions.

The definition of answer resolution in \SLGO{} (and SLG) differs from
resolution in Horn rules in order to take into account delay literals
in conditional answers.
\begin{definition}\label{def:ans-res2} 
Let $N$ be a node $A \mif D | L_1, ..., L_n,$ where $n>0$. Let $Ans
=A' \mif D' |$ be an answer whose variables are disjoint from $N$.
$N$ is {\em \SLGO{} resolvable} with $Ans$ if $\exists i$, $1 \leq i \leq
n$, such that $L_i$ and $A'$ are unifiable with an mgu\footnote{most
general unifier} $\theta$.  The \SLGO{} resolvent of $N$ and $Ans$ on
$L_i$ has the form:
\[ (A \mif D | L_1, ..., L_{i-1}, L_{i+1}, ..., L_n)\theta \]
if $D'$ is empty; otherwise the resolvent has the form:
\[ (A \mif D, L{_i}| L_1, ..., L_{i-1}, L_{i+1}, ..., L_n)\theta \] 
\end{definition}
Note that this form of resolution delays $L{_i}$ rather than
propagating the answer's delay list $D'$.  This is necessary, as shown
in \cite{CheW96}, to ensure polynomial data complexity.\footnote{If
delay lists were propagated directly, then delay lists could contain
all derivations which could be exponentially many in the worst case.}

Next, we relate different types of literals to their underlying subgoals.

\begin{definition}\label{def:underlying-subgoal}
The \emph{underlying subgoal} of $L$ is 1) $L$ if $L$ is a positive
literal or $L=\neg S$; 2) is $S$ if $L = \nf S$ (and $S$ is not based
on one of the new predicates $NH$ introduced in
Definition~\ref{d:MKNF-cohtransformKd}); or 3) is $\neg H(\vec{t_H})$ if
$L=\nf NH(\vec{t_H})$.  
\end{definition}

The first set of operations that we present deals with the creation of new trees and with resolution with program rules and with answers in other trees. 

\begin{definition} [SLG($\cO$) Operations -- 1] \label{def:cass-ops-1}
Let $\mathcal{K}^d=(\mathcal{O},\cO^d,\mathcal{P}^d)$ be a doubled Hybrid MKNF knowledge base. 
Further assume that a fixed selection function is used to select a literal from the $Goals$ in a node.

Given a forest $\cF_n$ of an SLG($\cO$) evaluation of $\mathcal{K}^d$, $\cF_{n+1}$ may be produced by one of the following operations.

\begin{enumerate}
\item  \subgc:
Let $\cF_n$ contain a tree with non-root node 
\[
N = Ans \mif Delays | G, Goals
\]
where $S$ is the underlying subgoal of $G$.
Assume $\cF_n$ contains no tree with root $S$.  Then add the tree
$S \mif | S$ to $\cF_n$.
\item \pgmcr:
Let $\cF_n$ contain a tree with root node $N = S \mif | S$ and $C$ be a rule
$Head \mif Body$ such that $Head$ unifies with $S$ with mgu
$\theta$.  Assume that in $\cF_n$, $N$ does not have a child
$N_{child} = (S \mif | Body)\theta$.  Then add $N_{child}$ as a child
of $N$.
\item \anscr:
Let $\cF_n$ contain a tree with non-root node $N$ whose selected
  literal $S$ is positive.  Let $Ans$ be an answer for $S$ in
  $\cF_n$ and $N_{child}$ be the \SLGO{} resolvent of $N$ and $Ans$ on
  $S$.  Assume that in $\cF_n$, $N$ does not have a child $N_{child}$.
  Then add $N_{child}$ as a child of $N$.
\end{enumerate}
\end{definition}

As illustrated in Example~\ref{e:wf2}, the
operation \subgc{} creates a new tree in the forest $\cF$ for a
selected literal in the {\em Goals} of some (non-root) node in a tree
in $\cF$.
Once a root node $N$ for a positive literal is created, the \pgmcr{} operation can
create children for $N$, given the rules in the knowledge base.
\anscr{} resolves positive literals
in nodes, with answers already in the forest, according to
Definition \ref{def:ans-res2}.
Contrary to SLG, the \subgc{} operation may also create new trees 
for classically negated literals to which only the operation \orcr{},
defined below, applies.

Now, if a sequence of \SLGO{} operations yields a (possibly intermediate)
forest containing an unconditional answer, then this answer is considered to be true.
Likewise, if no more operations are applicable to a set of trees, and none of them
contains an unconditional answer, i.e., the set of literals associated to these trees
is completely evaluated (see Definition~\ref{def:comp-eval}),
then we can interpret all these literals as false.
Expanding on this
correspondence, we may associate an \SLGO{} forest with a partial
interpretation, taking into consideration that, besides atoms and
default negated atoms, \SLGO{} also allows classically
negated literals as the roots of trees.  This interpretation is shown
to correspond to $M_{WF}$ (cf.\ Theorem~\ref{th:correct} below).

\begin{definition} \label{def:interp}
Let $\mathcal{F}$ be a forest.
Then the \emph{interpretation induced by $\mathcal{F}$}, $I_{\mathcal{F}}$, is the smallest set such that:
\begin{itemize}
\item A (ground) atom $A \in I_{\cF}$ iff $A$ is in the ground instantiation of some unconditional answer $Ans \mif |$ in $\cF$.
\item A (ground) negated atom $\neg A \in I_{\cF}$ iff $\neg A$ is in the ground instantiation of some unconditional answer $Ans \mif |$ in $\cF$.
\item A (ground) literal $\nf A \in I_{\cF}$ iff $A$ is in the ground instantiation of a literal whose tree in $\cF$ is marked as complete, and $A$ is not in the ground instantiation of any answer in a tree in $\cF$.
\end{itemize}  
An atom $S$ is \emph{successful} (resp. \emph{failed}) in $\cF$ if
$S'$ (resp. $\nf S'$) is in $I_\cF$ for every $S'$ in the ground
instantiation of $S$.  
\end{definition}

Whenever an atom $A$ is successful, we can fail its default negation $\nf A$. 
If an atom $A$ is failed, then we can simplify away $\nf A$. 
Ground default negated literals that are neither failed nor successful may be delayed and be simplified later.
More precisely:

\begin{definition} [SLG($\cO$) Operations -- 2] \label{def:cass-ops-2}
Let $\mathcal{K}^d=(\mathcal{O},\cO^d,\mathcal{P}^d)$ be a doubled Hybrid MKNF knowledge base, and assume a selection function as in Definition~\ref{def:cass-ops-1}.

Given a forest $\cF_n$ of an SLG($\cO$) evaluation of $\mathcal{K}^d$, $\cF_{n+1}$ may be further produced by one of the following operations.

\begin{enumerate}
  \setcounter{enumi}{3}
\item \negret:
Let $\cF_n$ contain a tree with a leaf node, whose selected literal $\nf S$ is ground
\[
N = Ans \mif Delays | \nf S, Goals.
\]
\begin{enumerate}
\item \negsuc:
If $S$ is failed in ${\cF_n}$ then create a child for $N$ of the form:
$
Ans \mif Delays | Goals.
$
\item \negfail:
If $S$ succeeds in ${\cF_n}$, then create a child for $N$ of the form
\emph{fail}.
\end{enumerate}
\item \delay{}:
Let $\cF_n$ contain a tree with leaf node $$N = Ans \mif Delays | \nf S,
Goals$$ such that $S$ is ground in $\cF_n$, but $S$ is neither
successful nor failed in $\cF_n$.
Then create a child for $N$ of the form $Ans$ $\mif Delays, \nf S|
Goals$.
\item \simpl{}:
Let $\cF_n$ contain a tree with leaf node $$N = Ans \mif Delays |$$
and let $L \in Delays$
\begin{enumerate}
\item If $L$ is failed in $\cF$ then create a child {\emph fail} for
$N$. 
\item If $L$ is successful in $\cF$, then create a child
$Ans \mif Delays' |$ for $N$, where $Delays'$ $=$ $Delays - L$.
\end{enumerate}
\end{enumerate}
\end{definition}

In Hybrid MKNF knowledge bases, an atom $S$ is true if it is derivable from the rules or from the DL part of the knowledge base. 
So far, we have presented the operation \pgmcr{} that handles the former case. 
We now introduce the \orcr{} operation to deal with the latter.

The next definition characterizes the behavior of an
abstract oracle, $\cO$,\footnote{We overload $\cO$ syntactically to
represent the oracle and the ontology, i.e., its underlying DL
knowledge base, since from the viewpoint of \SLGO{} they are the
same.} that computes entailment according to the DL knowledge base
$\cO$, to be used in the \orcr{} operation.  For that purpose, we define an oracle transition function that, given an interpretation induced by a forest, computes in a single step all possible atoms required to prove a goal $S$.  In other words, such an oracle, when presented with $S$ and a forest $\cF$,
non-deterministically returns in one step a set of ground atoms
$\cL$
such that: for each $L \in \cL$ there is at least one rule with $L$ in the head in ground $\cP_G$, and if $\cL$ were added to
$\cO$ augmented with $I_\cF$, the extended theory would immediately entail $S$.  We only have
to take into account that we appropriately query $\cO$ or its renaming
$\cO^d$ in a doubled Hybrid MKNF knowledge base, and that we extend $\cO$ only with the positive part of $I_{\mathcal{F}}$.

\begin{definition}\label{d:comptrans}
Let $\mathcal{K}^d=(\mathcal{O},\cO^d,\mathcal{P}^d)$ be a doubled Hybrid MKNF knowledge base, $S$ a ground goal, $\cL$ a set of ground atoms such that each $L\in\cL$ is unifiable with at least one rule head in $\mathcal{P}^d$, and $I_{\cF}^+ = I_{\cF}\setminus\{\nf A\mid\nf A\in  I_{\cF}\}$.
The \emph{complete oracle} for $\cO$, denoted $compT_{\cO}$, is defined by
\[
compT_{\cO}(I_{\cF},S,\cL) \text{ iff } \cO \cup I_{\cF}^+ \cup \cL  \models S \text{ or } \cO^d \cup I_{\cF}^+ \cup \cL \models S
\]
\end{definition}

\begin{example}
Consider the Hybrid MKNF knowledge base $\mathcal{K}$ containing $\cO$.
\begin{align*}
\tt{C(a)} & & \tt{C} \sqcap F \sqsubseteq \tt{E}
\end{align*}
Assume that $I_{\cF}$ is empty, and that there is at least one rule whose head unifies with $\tt{F(a)}$.
We query for $\tt{E(a)}$.  
In this case, $compT_{\cO}(\emptyset,\tt{E(a)},\{\tt{F(a)}\})$ holds because $\cO\cup\{\tt{F(a)}\}\models \tt{E(a)}$.
Thus, deriving $\tt{F(a)}$ from the rules would be enough to conclude that $\tt{E(a)}$ is true in the well-founded MKNF model.
\end{example}

The set $\cO \cup I_{\cF}^+ \cup \cL$ (and likewise $\cO^d \cup I_{\cF}^+ \cup \cL$) may be inconsistent even though the well-founded MKNF model of $\mathcal{K}$ exists.
Consequently, such a complete oracle potentially allows us to obtain a large number of entailments that are eventually useless to derive $S$ if $\cK$ is MKNF-consistent.
\begin{example}
Consider the Hybrid MKNF knowledge base $\mathcal{K}$ containing $\cO$.
\begin{align*}
\tt{C(a)} & & \tt{C} &\sqsubseteq \neg \tt{D} & \tt{E} &\sqsubseteq \tt{F} 
\end{align*}
Assume that $I_{\cF}$ is empty and we query for $\tt{E(a)}$.  If $\cO$
were extended with $\tt{D(a)}$, $\cO$ would become inconsistent, so
that all statements would be derivable from the extended $\cO$,
including $\tt{E(a)}$.  Hence
$compT_{\cO}(\emptyset,\tt{E(a)},\{\tt{D(a)}\})$ holds because
$\cO\cup\{\tt{D(a)}\}$ is inconsistent.  However, as $\cK$ is
MKNF-consistent, $\tt{D(a)}$ cannot be derived so that the
corresponding tree eventually fails.  In Section \ref{s:prop}, we
provide the definition of a partial oracle which overcomes this lack
of efficiency, and upon which concrete oracles can be based.
\end{example}

Complete oracles are applied to define the next \SLGO{} operation, which has no correspondence in SLG: 

\begin{definition} [SLG($\cO$) Operations -- 3] \label{def:cass-ops-3}
Let $\mathcal{K}^d=(\mathcal{O},\cO^d,\mathcal{P}^d)$ be a doubled Hybrid MKNF knowledge base.
Given a forest $\cF_n$ of an SLG($\cO$) evaluation of $\mathcal{K}^d$, $\cF_{n+1}$ may be produced by:
\begin{enumerate}
  \setcounter{enumi}{6}
\item \orcr:
Let $\cF_n$ contain a tree with root node $N = S \mif | S$, and suppose that $compT_{\cO}(I_{\cF_n},S,Goals)$ holds.
Assume that $N$ does not have a child $N_{child} = S \mif | Goals$ in $\cF_n$.\footnote{For that comparison, we consider the sequences $Goals$ as sets to avoid that one root node has several children whose sequences $Goals$ are merely permutations.}
Then add $N_{child}$ as a child of $N$.
\end{enumerate}
\end{definition}

\SLGO{} also includes an operation that marks a set of trees as complete if the corresponding set of literals is completely evaluated.
Completed trees can be used in \SLGO{} to simplify other trees and to augment the interpretation associated with the forest with default negated literals

\begin{definition}\label{def:comp-eval}
A set $\mathcal{S}$ of literals in a forest $\mathcal{F}$
is \emph{completely evaluated} if at least one of the conditions holds
for each $S \in {\mathcal S}$
\begin{enumerate}
\item The tree for $S$ contains an answer $S \mif |$; or
\item For each node $N$ in the tree for $S$:
\begin{enumerate}
\item The underlying subgoal of the selected literal of $N$ is marked as complete; or
\item The underlying subgoal of the selected literal of $N$ is in $\mathcal{S}$ and there are no applicable \subgc, \pgmcr,  
\anscr{} (Definition~\ref{def:cass-ops-1}), \negret, \delay{} (Definition~\ref{def:cass-ops-2}) or \orcr{} (Definition~\ref{def:cass-ops-3}) operations for $N$.
\end{enumerate}
\end{enumerate}
\end{definition}
Once a set of literals is determined to be completely evaluated,
a \compl{} operation marks the trees for each literal
(Definition~\ref{def:forest}).  If a subgoal $S$ is completed due to
condition 1 holding, we say that $S$ is {\em early completed}.  If
condition 1 does not hold, condition 2a of the above definition
prevents the \compl{} operation from being applied to one of a set of
trees if certain other operations are applicable to those
trees.
This notion
of completion is incremental in the sense that once a set $\cS$ of
mutually dependent subgoals is fully evaluated, the derivation need
not be concerned with the trees for $\cS$ apart from any answers they
contain.  In an actual implementation resources for such trees can be
reclaimed.

In certain cases the propagation of conditional answers through
resolution (Definition \ref{def:ans-res2}) can lead to a set
of \emph{unsupported answers} --- conditional answers that are false
in the well founded model (see, e.g., Example 1
of \cite{SPP09}).\footnote{As an aside, we note that unsupported
answers appear to be uncommon in practical evaluations which minimize
the use of delay such as~\cite{SaSW99}.}  Intuitively, these answers,
which have positive mutual dependencies, correspond to an unfounded
set, but their technical definition is based on the form of
conditional answers.

\begin{definition}\label{def:sup-ans2}
Let $\mathcal{F}$ be an \SLGO{} forest, and $Answer$ be an atom that
occurs in the head of some answer in a tree with root $S$. Then
$Answer$ is supported in $\mathcal{F}$ if and only if:
\begin{enumerate}
\item $S$ is not completely evaluated; or
\item there exists an answer node $Answer' \mif Delays |$ in $S$ such
that $Answer'$ subsumes $Answer$ and for every positive literal $L \in
Delays$, $L$ is supported in $\mathcal{F}$.
\end{enumerate}
\end{definition}

We are now able to characterize the last two operations of \SLGO{}: one allows the completion of trees, and the other removes unsupported answers.  

\begin{definition} [SLG($\cO$) Operations -- 4] \label{def:cass-ops-4}
Let $\mathcal{K}^d=(\mathcal{O},\cO^d,\mathcal{P}^d)$ be a doubled Hybrid MKNF knowledge base.
Given a forest $\cF_n$ of an SLG($\cO$) evaluation of $\mathcal{K}^d$, $\cF_{n+1}$ may also be produced by one of the following operations.
\begin{enumerate}
  \setcounter{enumi}{7}
\item \compl:
Given a completely evaluated set $\cal{S}$ of literals
(Definition~\ref{def:comp-eval}), mark the trees for all literals in
$\cal{S}$ as complete.
\item \anscompl: \label{last-operation}
Given a set of unsupported answers $\cal{UA}$, 
create a failure node as a child for each answer $Ans \in \cal{UA}$.
\end{enumerate}
\end{definition}

Each of the operations (1)--(\ref{last-operation}), in Definitions~\ref{def:cass-ops-1}, \ref{def:cass-ops-2}, \ref{def:cass-ops-3} and \ref{def:cass-ops-4}, can be seen as a
function that associates a forest with a new forest by adding a new
tree, adding a new node to an existing tree, or marking a set of trees
as complete. The only thing missing to complete the description of
the procedure is the formalization of the initialization of an \SLGO{}
evaluation, i.e., how the initial (DL-safe) conjunctive query is
defined.

\begin{definition}
Let $\mathcal{K}^d$ be a doubled Hybrid MKNF knowledge base and let
$q$ be a query of the form $q(X_{i})\leftarrow A_{1}, \ldots,
A_{n}, \nf B^{d}_{1}, \ldots, \nf B^{d}_{m}$ where $X_{i}$ is the
(possibly empty) set of requested variables.  We set $\mathcal{F}_{0}
= \{q(X_{i})\mif\mid q(X_{i})\}$ to be the initial forest of an
\SLGO{} evaluation of $\mathcal{K}^{d}$ for $q$ and add $q$ itself
to $\mathcal{K}^d$.
\end{definition}
Of course, if the query is atomic we can simply start with the query
itself, i.e., with the root containing the queried literal itself.
Since the derivation uses $\mathcal{K}^{d}$ (the doubled knowledge
base), the technically correct way to query negative literals is to
use $\nf B^{d}$ instead of $\nf B$ for any atom $B$ which is why we
use the doubled predicates for negative literals in the query.

Finally, note that if $\cO$ represents an expressive DL, then $\cO$
may derive equalities between different individuals because the unique
names assumption (UNA) is not applied.  Hybrid MKNF accounts for that
using the standard names assumption (see \cite{MR10,KAH:AI11}), thus
adapting reasoning with equalities as well.  As such, equalities allow
us to derive further information in the sense that, for example, if
$\tt{C(a)}$ and $\tt{a}\approx \tt{b}$ hold, then $\tt{C(b)}$ is
derivable.  If $\tt{C(a)}$ is a DL-atom, then the DL reasoner of the
oracle takes care of the problem internally.  Only if $\tt{C(a)}$ is a
non-DL-atom, then we specifically have to query for equalities in
$\cO$ which is why \orcr{} is not restricted to DL-atoms.

In the next section, we show that \SLGO{} always terminates (Theorem~\ref{t:terminator}) and, even though some orders of application of the possible operations are more efficient than others, that the procedure is confluent (Theorem~\ref{t:appord}).
We also show that the procedure is sound and complete w.r.t.\ the well-founded MKNF model (Theorem~\ref{th:correct}) and that it is sound w.r.t.\ the semantics of two-valued MKNF (Corollary~\ref{c:2vMKNFSound}).
Finally, under some assumptions, we maintain the computational complexity of the bottom-up procedure (Theorem~\ref{t:dcpo}), which is actually an improvement since we do not have to consider the entire knowledge base but only the part relevant for a concrete query. 
But before showing these results, we finish the presentation
of \SLGO{} with an example illustrating its behavior.
\begin{example}

In order to illustrate the actions of \SLGO{} we consider a derivation
of an answer to the query {\tt discount(Bill)} using a KB $\cK$ from
\cite{MR07}:\footnote{For ease of reading and since neither an
  MKNF-inconsistency nor an issue related to coherence occurs, we
  operate on $\mathcal{K}$ directly instead of on $\mathcal{K}^{d}$.}

\begin{align}
\tt{NonMarried} \equiv & \,\tt{\neg Married} \label{e:81}\\
\neg \tt{Married}  \sqsubseteq &\, \tt{HighRisk} \label{e:82}\\
\exists \tt{Spouse}.T  \sqsubseteq &\, \tt{Married} \label{e:83}\\
& \,(\exists \tt{Spouse}.\{\tt{Michelle}\})(\tt{Bill}) \label{e:84}\\
\tt{NonMarried}(\tt{x})  \leftarrow &\, \nf \tt{Married}(\tt{x})\label{e:85}\\
\tt{discount}(\tt{x})  \leftarrow &\,\nf \tt{HighRisk}(\tt{x})\label{e:86}
\end{align}

First, note that TBox and ABox information are each distributed over
both the DL KB and the rules.  Figure~\ref{fig:forest}
shows the final forest for this evaluation, where elements are marked
in the order they are created.  The initial forest for the evaluation
consists of node 1 only.  Given the selected literal of node 1,
$\tt{discount}(\tt{Bill})$, we can only apply \pgmcr{}, so we use rule
(\ref{e:86}) to produce node 2, followed by \newsg{} to produce node
3.  No rules are applicable for node 3,
$\tt{HighRisk}(\tt{Bill})$, but an \orcr{} operation can be applied to
derive from axioms (\ref{e:81}) and (\ref{e:82}) that if
$\tt{NonMarried}(\tt{Bill})$ can be proven (node 4), then this
suffices to prove $\tt{HighRisk}(\tt{Bill})$.  Then, via a \newsg{}
operation, node 5 is obtained.  For the selected literal in node 5,
$\tt{NonMarried}(\tt{Bill})$, \pgmcr{} produces node 6
from (\ref{e:85}) and \newsg{} produces node 7.  The selected literal
of node 7, $\tt{Married}(\tt{Bill})$, is not the
head of a rule, so the only possibility is to use
\orcr{}, and the answer $\tt{Married}(\tt{Bill})$ is derived from
axioms (\ref{e:83}) and (\ref{e:84}).  Using this answer, the tree for
$\tt{Married}(\tt{Bill})$ can be early completed and a \negret{}
operation produces node 10.  The tree for $\tt{NonMarried}(\tt{Bill})$,
which does not have an answer, must be completed (step 11), and the
same holds for $\tt{HighRisk}(\tt{Bill})$ (step 12).  Once this
occurs, a \negret{} operation is enabled to produce node 13.

\begin{figure}[t]
\includegraphics[width=14cm]{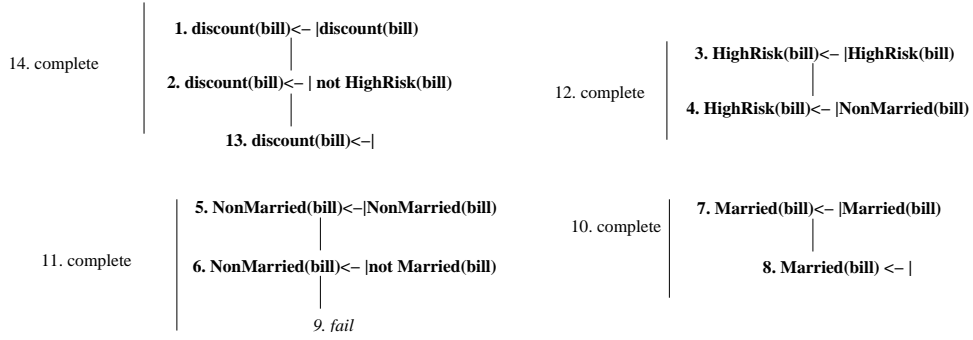}
\caption{Final Forest for the query $\tt{discount}(\tt{Bill})$ to $\cK$.}\label{fig:forest}
\end{figure} 
\end{example}

The evaluation in the example illustrates two main points.
First, the evaluation makes use of classical negation in the ontology along with closed world negation in the rules.
Second, the actions of the DL part and the program part are interleaved, with the program ``calling'' the oracle by \orcr{}, and the oracle ``calling'' the program back with the answers of that operation.

\section{Properties of Tabled \SLGO{}-resolution}\label{s:prop}

We now present several properties of \SLGO{}-resolution.
The first property we can ensure is that our extension of SLG resolution terminates for the evaluation of any query, generating a final forest.

\begin{theorem}\label{t:terminator}
Let $q = L$ be a query to a doubled Hybrid MKNF knowledge base $\cK^d$.
Then any \SLGO{} evaluation of $\K^d$ for $q$ terminates after finitely many steps, producing a finite \emph{final forest}.
\end{theorem}
\begin{proof}
The proof is straightforward since we know already that SLG, i.e., \SLGO{} without \orcr{} and the extended \newsg{} operation, terminates finitely for programs with bounded term-depth, and transfinitely otherwise (cf.\ Theorem 5.10 of \cite{CheW96}).
Since Definition~\ref{d:MKNFKB} ensures that Hybrid MKNF knowledge bases do not contain recursive terms, i.e., non-nullary functors, they have bounded term depth, and so does the doubled knowledge base $\cK^d$.
Accordingly, we only have to ensure that the new operation \orcr{} and the extension of \newsg{} do not invalidate finite termination.

The operation \orcr{} can be applied in the same situation as \pgmcr{}, namely when creating a new child for a root of a tree, so that each operation can be applied only once to a given node (for each of the finitely many rules, respectively for each of the finitely many possible answers of the complete oracle), and this creates one child per successful application.
Now, since the knowledge base $\cK^d$ is finite, the number of (ground) rule heads is finite.
Thus, 1) the number of children possibly created with \orcr{} for any arbitrary root is finite; and 2) the size of the nodes created is also finite.

The extension of the operation \newsg{} creates even in the worst case
finitely many more trees with roots to which only \orcr{} is
applicable, which in its turn is fintely many, as just demonstrated.

We conclude that termination holds for \SLGO{}.
\end{proof}

As \SLGO{} is defined, there is no prescribed order in which to apply the operations possible in a forest $\cF_{i}$.
For SLG some orders of application are in general more efficient than others but, as shown in \cite{CheW96}, any order yields the same outcome for any query.
This same sort of confluence also holds for \SLGO{}:

\begin{theorem}\label{t:appord}
Let $\cE_1$ and $\cE_2$ be two \SLGO{} evaluations of a query $q = L$ to a doubled Hybrid MKNF knowledge base $\mathcal{K}^{d}$, $\cF_1$ the final forest of $\cE_1$, and $\cF_2$ the final forest of $\cE_2$.
Then, $I_{\cF_1} = I_{\cF_2}$.
\end{theorem}
\begin{proof}

This is a well-known property for SLG as defined using the operations of Definition~\ref{def:cass-ops-1} excluding the extension of \newsg{} to classical negation, and the operations of Definitions \ref{def:cass-ops-2} and \ref{def:cass-ops-4} (cf.\ Theorem 5.7 of~\cite{CheW96}).
Accordingly, we consider cases in which $\cE_1$ and $\cE_2$ make use of the operations that have been introduced/extended in \SLGO{}.
However, \pgmcr{} is used in SLG, and if we just consider the created children, then \pgmcr{} and \orcr{} are not distinguishable.
Thus, we can consider that \orcr{} is a syntactic variant of \pgmcr{}.
The same holds for \newsg{} and the treatment of default negated atoms $\nf S$ that create a tree with root $S$ and those special literals $\nf NH(t_i)$ that may allow us to create a tree with root $\neg H(t_i)$: both its children are not distinguishable and only one of the two is applicable in each case.
Thus, confluence of \SLGO{} follows directly from confluence of SLG (see Theorem 5.7 of~\cite{CheW96}).
\end{proof}

The above theorem is also helpful to prove that \SLGO{} is a correct
query procedure for \MKNF{} and terminates within the same complexity
bounds as the semantics defined in \cite{KAH:AI11}.  First, we show
that \SLGO{} coincides with \MKNF{}.  Intuitively, what we have to
show is that the well-founded MKNF model, as presented in
Section \ref{s:prel} and based on the computation presented in
Section \ref{s:altcomp}, coincides with the interpretation
$I_{\mathcal{F}}$ induced by the final forest $\mathcal{F}_{n}$ for some query $q$ to
$\mathcal{K}^{d}$ in all the ground literals involved in the query.\footnote{Without loss of generality, we can restrict that statement to ground queries, a non-ground query would simply require to check all the ground instances.}
We can further simplify that by showing, for each literal $L$ appearing in
$\mathcal{K}^{d}_{G}$, that $L\in M_{WF}$ (Definition~\ref{d:MKNFWFS})
if and only if $L\in I_{\mathcal{F}}$ with ground query $q= {L}$ and
$\mathcal{F}_{n}$ for some $n$.  Note that this correspondence also
holds for atoms and classically negated atoms only appearing in the
ontology.

\begin{theorem}\label{th:correct}
Let $\mathcal{K}^d$ be a doubled Hybrid MKNF knowledge base, and $I_{\mathcal{F}}$ the interpretation induced by the final forest $\cF$ of an \SLGO{} evaluation of $\mathcal{K}^{d}$ for a ground query $q= {L}$ where $L$ is a literal or a classically negated atom.
\SLGO\ resolution is sound and complete w.r.t.\ $M_{WF}$, which is obtained from $\mathbf{P}_\omega^d$ and $\mathbf{N}_\omega^d$, i.e.,
\begin{itemize}
\item for $L\in {\sf KA}(\cK^d_G)$: 
\begin{itemize}
\item $L\in \mathbf{P}_\omega^d$ if and only if $L\in I_{\mathcal{F}}$ and 
\item $L_1^d\not\in \mathbf{N}_\omega^d$ if and only if $L=\nf L_1^d\in I_{\mathcal{F}}$.
\end{itemize}
\item for $L\not\in {\sf KA}(\cK^d_G)$: $\cO\cup \mathbf{P}_\omega^d\models L$ or $\cO^d\cup \mathbf{P}_\omega^d\models L$ if and only if $L\in I_{\mathcal{F}}$.  
\end{itemize}
\end{theorem}
\begin{proof}

\noindent \textbf{(Completeness):} 
We show by induction on $n$ that
\begin{itemize}
\item for $L\in {\sf KA}(\cK^d_G)$, if $L$ is a positive literal, then
  $L\in\mathbf{P}^{d}_{n}$ implies that $L\in I_{\mathcal{F}}$; and if
  $L=\nf L_1^d$ is a negative literal, then
  $L_1^d\not\in\mathbf{N}^{d}_{n}$ implies that $\nf L^d_1\in
  I_{\mathcal{F}}$
\item for $L\not\in {\sf KA}(\cK^d_G)$, if $\mathcal{O}\cup
  \mathbf{P}^{d}_{n}\models L$ or $\mathcal{O}^d\cup
  \mathbf{P}^{d}_{n}\models L$ then $L \in I_{\mathcal{F}}$.
\end{itemize}

The induction base holds immediately, for $L\in {\sf KA}(\cK^d_G)$, since $\mathbf{P}^{d}_{0}$ is empty and $\mathbf{N}^{d}_{0}$ contains all literals appearing in ${\sf KA}(\mathcal{K}^{d}_G)$.
For $L\not\in {\sf KA}(\cK^d_G)$, we obtain that $\mathcal{O}\models L$ or $\mathcal{O}^d\models L$, so we can create a tree $L:-\mid L$ and with \orcr{} an answer $L:-\mid$, which shows $L \in I_{\mathcal{F}}$.

({\em Induction Hypothesis $1$}) Now suppose the claim holds for $n$. We
have to show the induction step for $n+1$.  For $L\in {\sf
  KA}(\cK^d_G)$, let $L$ be a positive literal, and suppose that
$L\in\mathbf{P}^{d}_{n+1}$ but $L\not\in\mathbf{P}^{d}_{n}$ (otherwise
the claim would immediately follow by the induction hypothesis).
Therefore, $L\in \Gamma_{\mathcal{K}^d_G}(\mathbf{N}^{d}_{n})$ and so
$L\in T_{\mathcal{K}^{d}_G//'\mathbf{N}^{d}_{n}}\uparrow \omega$
(Definition~\ref{d:GammaKd}).  We show by induction on $m$
that $L\in T_{\mathcal{K}^{d}_G//'\mathbf{N}^{d}_{n}}\uparrow m$ implies
that $L\in I_{\mathcal{F}}$.

\noindent
{\em Inner  Induction}

The base case is void since $T_{\mathcal{K}^{d}_G//'\mathbf{N}^{d}_{n}}\uparrow 0$ is empty.

({\em Induction Hypothesis $2$}) Suppose the property holds for
  $m$, we show it for $m+1$.  So, assume that $L\in
  T_{\mathcal{K}^{d}_G//'\mathbf{N}^{d}_{n}}\uparrow (m+1)$ but
  $L\not\in T_{\mathcal{K}^{d}_G//'\mathbf{N}^{d}_{n}}\uparrow m$
  (otherwise the property would immediately follow by the induction
  hypothesis ($2$)).  Then $L \in
  T_{\mathcal{K}^{d}_G//'\mathbf{N}^{d}_{n}}(T_{\mathcal{K}^{d}_G//'\mathbf{N}^{d}_{n}}\uparrow
  m)$ and either $L \in
  R_{\mathcal{K}^{d}_G//'\mathbf{N}^{d}_{n}}(T_{\mathcal{K}^{d}_G//'\mathbf{N}^{d}_{n}}\uparrow
  m)$ (i.e., $L$ is a consequence of rule deduction,
  Definition~\ref{d:opRkDkTkd}) or $L \in
  D_{\mathcal{K}^{d}_G//'\mathbf{N}^{d}_{n}}(T_{\mathcal{K}^{d}_G//'\mathbf{N}^{d}_{n}}\uparrow
  m)$ ($L$ is a consequence of deduction in the ontology).  
\mycomment{ In the first case, $\mathcal{K}^{d}//'\mathbf{N}^{d}_{n}$
  contains a rule of the form $L\leftarrow A_1,\ldots, A_n$ such that
  all $\K A_{i}\in T_{\mathcal{K}^{d}/\mathbf{N}^{d}_{n}}\uparrow m$.
  Additionally, there is a rule $L\leftarrow A_1,\ldots, A_n,\nf
  B_1,\ldots, \nf B_m$ in $\mathcal{K}^{d}$ and all $B_{j}\not\in
  \mathbf{N}^{d}_{n}$.  }
In the first case, for $L$ to be the consequence of a rule derivation,
there must be a rule $L\leftarrow A_1,\ldots, A_n,\nf B_1,\ldots, \nf
B_m$ in $\mathcal{K}^{d}_G$ such that all $B_{j}\not\in
\mathbf{N}^{d}_{n}$ and all $\K A_{i}\in
T_{\mathcal{K}^{d}_G//'\mathbf{N}^{d}_{n}}\uparrow m$.  Such a rule gives
rise to the rule $L\leftarrow A_1,\ldots, A_n$ in the the MKNF-coherent
reduction, $\mathcal{K}^{d}_G//'\mathbf{N}^{d}_{n}$.
We thus know by the two induction hypotheses that all $A_{i}$ and all
$\nf B_{j}$ appear in $I_{\mathcal{F}}$.  From that we can construct a
tree with root $L:-\mid L$ and a child obtained by applying \pgmcr{}
with the rule considered.  In the resulting child the set of goals
contains exactly all $A_{i}$ that can be removed by \anscr{} and all
$\nf B_{j}$ that can be removed by \negret{}.  The result is a leaf
node $L:- \mid$ and we obtain that $L\in I_{\mathcal{F}}$ for this
order of applying \SLGO{} operations.  Since Theorem \ref{t:appord}
ensures that we achieve the same result if we alter the order of such
applications of \SLGO{} operations, we know that the statement holds in
general.  
In the second case, i.e., for $L \in
D_{\mathcal{K}^{d}_G//'\mathbf{N}^{d}_{n}}(T_{\mathcal{K}^{d}_G//'\mathbf{N}^{d}_{n}}\uparrow
m)$, we construct a tree $L:-\mid L$ and apply \orcr{} as (finitely)
many times as necessary.  One of those children is the one actually
allowing to derive $L$ by means of the ontology, 
i.e., all goals in this child are positive literals that are true in
$T_{\mathcal{K}^{d}_G//'\mathbf{N}^{d}_{n}}\uparrow m$.  We apply \anscr{}
to these literals, and this, by the induction hypothesis ($2$),
results in a leaf node $L:- \mid$.  As before, Theorem \ref{t:appord}
ensures that a different application order again yields eventually the
same result.

Now, let $L$ be a negative literal $\nf L_1^d$, and suppose that $L_1^d\not\in\mathbf{N}^{d}_{n+1}$ but $L_1^d\in\mathbf{N}^{d}_{n}$ (otherwise the claim would follow immediately by the induction hypothesis ($1$)).
Then, $L_1^d\in U_{\mathcal{K}^d_G}(\mathbf{P}^d_n,{\sf KA}(\mathcal{K}^d_G)\setminus\mathbf{N}^d_n)$ by Lemma~\ref{l:convAltWd}, i.e., $L_1^d$ occurs in the greatest unfounded set w.r.t.\ $(\mathbf{P}^d_n,{\sf KA}(\mathcal{K}^d_G)\setminus\mathbf{N}^d_n)$.
We construct a tree with root $L_1^d:- \mid L_1^d$.
We proceed by creating all children of that root, applying \pgmcr{} and \orcr{} as (finitely) many times as possible.
By Definition~\ref{d:unfMKNFd}, each such child (after finitely many subsequent operations) is either false or completely evaluated as in case 2.(b) of Definition~\ref{def:comp-eval}, which means that another element of $U_{\mathcal{K}^d_G}(\mathbf{P}^d_n,{\sf KA}(\mathcal{K}^d_G)\setminus\mathbf{N}^d_n)$ has been encountered in the list of goals.
Note that \SLGO{} selects literals in some order, while the greatest unfounded set $U$ just refers to some other element in $U$.
Consequently, it may happen that we have to evaluate some literals first whose evaluation is only known in an iteration step $m$ with $m>n$.
But this does not cause any problem.
Such negative literals may be simply delayed (\delay{}), while both positive and negative literals are processed (\subgc{} and so on): if a literal can eventually be resolved, then it is removed from the list of goals of a child.
Otherwise, we obtain an even larger unfounded set.
In both cases, once no further operation can be applied, the set $U$ can be completed, and, by Definition~\ref{def:interp},  we derive $\nf L_1^d\in I_\mathcal{F}$.

\noindent
{\em End of Inner Induction}

The previous inner induction handled the case where $L$ was derived as
a consequence of a rule.  Alternately, suppose that $L\not\in {\sf
  KA}(\cK^d_G)$ and $\mathcal{O}\cup \mathbf{P}^{d}_{n}\models L$ or
$\mathcal{O}^d\cup \mathbf{P}^{d}_{n}\models L$.  We can can construct
a tree starting with $L :-\mid L$ and apply \orcr{} until we get a
child $L:-\mid Goals$ such that $Goals\subseteq \mathbf{P}^{d}_{n}$,
which has to exist.  We apply \anscr{} to all positive literals in
$Goals$, which is possible by the induction hypothesis ($1$) thus
deriving the answer $L:-\mid$, from which we conclude $L \in
I_{\mathcal{F}}$.

\noindent \textbf{(Soundness):} 
We show by induction on $n$ that: 
\begin{itemize}
\item for $L\in {\sf KA}(\cK^d_G)$, if $L$ is a positive literal, then
  $L\in I_{\mathcal{F}_n}$ implies that $L\in\mathbf{P}^{d}_\omega$,
  and if $L=\nf L_{1}^d$ is a negative literal, then $L\in
  I_{\mathcal{F}_n}$ implies that
  $L_{1}^d\not\in\mathbf{N}^{d}_{\omega}$;  and
\item for $L\not\in {\sf KA}(\cK^d_G)$, if $L \in I_{\mathcal{F}}$, then
  $\mathcal{O}\cup \mathbf{P}^{d}_{\omega}\models L$ or
  $\mathcal{O}^d\cup \mathbf{P}^{d}_{\omega}\models L$.
\end{itemize}

The induction base holds trivially, since $I_{\mathcal{F}_0}$ is empty.
So assume the property holds for $n$.  We show that the property holds
for all cases (1)--(\ref{last-operation}) of an \SLGO{} operation
that may be applied to $I_{\mathcal{F}_{n}}$ yielding $I_{\mathcal{F}_{n+1}}$.
\begin{enumerate}
\item \subgc: This operation creates a new tree and does alone not alter $I_{\mathcal{F}}$, i.e., if $L\in I_{\mathcal{F}_{n+1}}$, then $L\in I_{\mathcal{F}_n}$, and the property holds by the induction hypothesis.
\item \pgmcr: A new child is created for the root $S \mif{} \mid S$. 
If this child has an empty list of goals, then a rule with empty body was used to create this child.
Now, if $L$ is a positive literal with $L=S$, then $L\in I_{\mathcal{F}_{n+1}}$.
But then, $L\in \mathbf{P}^{d}_\omega$ since there is a fact $L$ in $\mathcal{K}^d_G$.
Alternatively, if the list of children is not empty, then $L\in I_{\mathcal{F}_{n+1}}$ implies $L\in I_{\mathcal{F}_n}$, and the property holds by the induction hypothesis.
\item \anscr: If the resolved goal is the last remaining, then the outcome of the operation is an unconditional answer.
Suppose the answer template is equal to $L$.
We can trace back this child to the immediate child of the root.
All goals in this particular child have already been resolved, so that, by the induction hypothesis, all positive literals $L$ appear in $\mathbf{P}^{d}_\omega$ and all negative literals $\nf L_{1}^d$ do not appear in $\mathbf{N}^{d}_{\omega}$.
But then the property holds, no matter whether  $L\in {\sf KA}(\cK^d_G)$ or not.
Alternatively, if the list of goals (including delayed ones) is not empty, then $L\in I_{\mathcal{F}_{n+1}}$ implies $L\in I_{\mathcal{F}_n}$, and the property holds by the induction hypothesis.
\item \negret
\begin{enumerate}
\item \negsuc: The argument is exactly the same as for \anscr{}, only now the last goal is a negative literal.
\item \negfail: This operation fails one child. 
However, it does alone not contribute to $I_{\mathcal{F}}$, i.e., if $L\in I_{\mathcal{F}_{n+1}}$, then $L\in I_{\mathcal{F}_n}$, and the property holds by the induction hypothesis.
\end{enumerate}
\item \delay: This operation at best provides a conditional answer.
As such it does not affect $I_{\mathcal{F}}$ alone.
Therefore, if $L\in I_{\mathcal{F}_{n+1}}$, then $L\in I_{\mathcal{F}_n}$, and the property holds by the induction hypothesis.
\item \simpl:
\begin{enumerate}
\item The first simplification case corresponds exactly to \negfail{},
  only here the failure occurs in {\em Delays} and the failed literal
  may be positive or negative.
\item The second simplification case corresponds exactly to \negsuc{},
  only now the success occurs in {\em Delays} and the successful
  literal may be positive or negative.
\end{enumerate}
\item \orcr:  A new child is created for the root $S\mif{} \mid S$ by means of the oracle.
If the returned list of goals is empty, then the oracle allows us to derive the root directly, and $\mathcal{O}\cup I_{\mathcal{F}_n}\models L$ or $\mathcal{O}^d\cup I_{\mathcal{F}_n}\models L$.
In this case, if $L$ is a positive literal with $L=S$, then $L\in I_{\mathcal{F}_{n+1}}$.
If $L\in {\sf KA}(\cK^d_G)$, then $L\in \mathbf{P}^{d}_\omega$, since the operator $D_{\mathcal{K}^d_G}$ together with all $L'\in I_{\mathcal{F}_n}$, for which $L'\in \mathbf{P}^{d}_\omega$ holds by the induction hypothesis, allows us to derive $L$.
If $L\not\in {\sf KA}(\cK^d_G)$, then $\mathcal{O}\cup \mathbf{P}^{d}_{\omega}\models L$ or $\mathcal{O}^d\cup \mathbf{P}^{d}_{\omega}\models L$ holds since we have that, for all $L'\in I_{\mathcal{F}_n}$, $L'\in \mathbf{P}^{d}_\omega$ holds by the induction hypothesis.
Alternatively, if the list of goals is not empty, then $L\in I_{\mathcal{F}_{n+1}}$ implies $L\in I_{\mathcal{F}_n}$, and the property holds by the induction hypothesis.
\item \compl: This operation only affects $I_{\mathcal{F}}$ if some $A$ is in the ground instantiation of a completely evaluated literal in $\mathcal{F}$ and $A$ is not in the ground instantiation of any answer in a tree in $\mathcal{F}$.
In other words, this operation introduces $\nf L'$ to $I_\mathcal{F}$.
In particular, consider $L=\nf L_{1}^d$ as a negative literal and $L\in  I_{\mathcal{F}_{n+1}}$.
Thus, the tree for $L_1^d$ does not contain any answer but also no further operation can be applied, i.e., in each child, there is (at least) one literal that either can not be resolved or it is failed.
This matches the condition of the greatest unfounded set and we obtain that $L_1^d\not\in \mathbf{N}^{d}_{\omega}$. 
For all other cases, if $L\in I_{\mathcal{F}_{n+1}}$, then $L\in I_{\mathcal{F}_n}$, and the property holds by the induction hypothesis.
\item \anscompl: This operation may affect $I_{\mathcal{F}}$ by adding
  failure nodes as children to conditional answers.  Assume that one
  such answer occurs within some tree with root $S \mif{} S$ in
  $\cF_n$.  In such a case, $S$ may become false in $I_{\cF_{n+1}}$ but
  was not false in $I_{\cF_n}$.  However, the notion of an answer that is
  not supported (Definition~\ref{def:sup-ans2}) captures the
  definition of an element of an unfounded set: in this case
  literals in the unfounded set may be in the {\em Delays} of an
  answer. As with the case of \compl{}, we have that for any $L$ in
  the ground instantiation of $S$, $L^d\not\in
  \mathbf{N}^{d}_{\omega}$.  For all other cases, if $L\in I_{\mathcal{F}_{n+1}}$, then $L\in I_{\mathcal{F}_n}$, and the property holds by the induction hypothesis.

\end{enumerate}
We conclude that soundness holds.
\end{proof}

Given the soundness of \MKNF{} with respect to the semantics of MKNF
knowledge bases of \cite{MR10}, it follows easily from \cite{KAH:AI11} that:

\begin{corollary}\label{c:2vMKNFSound}
Let $\mathcal{K}$ be a Hybrid MKNF knowledge base and $L$ a literal that appears in $\mathcal{K}^{d}_{G}$.
If $L\in I_{\mathcal{F}}$ ($L=\nf L_1^d\in I_{\mathcal{F}}$ respectively), where $I_{\mathcal{F}}$ is induced by the forest $\cF$ of an  \SLGO{} evaluation of $\mathcal{K}^{d}_{G}$ for query $q= {L}$, then $L$ ($\nf L_1$ respectively) is derivable from all two-valued MKNF models of $\mathcal{K}$.
\end{corollary}
In addition to the interpretation of the final forest $I_{\cF}$ being sound with respect to the 2-valued MKNF model, the conditional answers in $\cF$ can be seen as a well-founded reduct of the rules in $\cK$, augmented with conditional answers derived by \orcr{} operations.
As a result, the final forest can be seen as a {\em residual program}: a sound transformation not only of the rules, but of information from the oracle, and can be used to construct a partial 2-valued stable model.\footnote{\cite{CheW96} discusses these transformational aspects of SLG resolution, which are preserved in \SLGO{}, while the XSB manual discusses how the residual program can serve as input to an ASP solver.}

Regarding complexity, it is clear that the complexity of the whole procedure \SLGO{} depends on the complexity of the oracle, and also on the number of  results returned by each call to the oracle.
Clearly, the complexity associated to the computation of one result of the oracle function is a lower-bound of the complexity of \SLGO.
Moreover, even if the computation of one result of the oracle is tractable, the (data) complexity of \SLGO{}  may still be exponential if exponentially many solutions are generated by the oracle, e.g., if returning all supersets of a solution.
This is so, because our definition of the oracle is quite general, and in order to prove interesting complexity results some assumptions must be made about the oracle.
We start by defining a correct partial oracle:

\begin{definition}\label{d:parttrans}
Let $\mathcal{K}^d=(\mathcal{O},\cO^d,\mathcal{P}^d)$ be a doubled Hybrid MKNF knowledge base, $S$ a goal, and $\cL$ a set of ground atoms such that each $L\in \cL$ is unifiable with at least one rule head in $\mathcal{P}^d$ (called program atoms). 
A \emph{partial oracle} for $\cO$, denoted $pT_{\cO}$, is a relation $pT_{\cO}(I_{\cF},S,\cL)$ such that if $pT_{\cO}(I_{\cF},S,\cL)$, then
\begin{align*}
\cO \cup I_{\cF}^+ \cup \cL  &\models S  \text{ and } \cO \cup I_{\cF}^+\cup \cL\text{ consistent; or }\\
\cO^d \cup I_{\cF}^+ \cup \cL &\models S  \text{ and } \cO^d \cup I_{\cF}^+\cup \cL\text{ consistent.}
\end{align*}
A partial oracle $pT_{\cO}$ is \emph{correct} w.r.t.\ $compT_{\cO}$ iff, for all MKNF-consistent $\mathcal{K}^d$, replacing $compT_{\cO}$ in \SLGO{} with $pT_{\cO}$ succeeds for exactly the same set of queries. 
\end{definition}

Note that the complete oracle is indeed generating unnecessarily many answers, and it can be replaced by a partial one that assures correctness.
E.g., consider a partial oracle that does not return supersets of other results. Such a partial oracle is obviously correct.
A further improvement on efficiency is the restriction to consistent sets $\cO \cup I_{\cF_n}^+\cup L$ and $\cO^d \cup I_{\cF_n}^+\cup L$.
If the knowledge base is MKNF-consistent, then looking for derivations based on inconsistencies is pointless anyway: we would just create a potentially large number of children none of which would result in an unconditional answer. 
In this sense, partial oracles are limited to meaningful derivations.
In the case of an MKNF-inconsistent knowledge base, things get a bit more complicated.

\begin{example}\label{e:double2}
Consider again the already doubled knowledge base from Example~\ref{e:double}.
\begin{align*}
\tt{Q}  & \sqsubseteq  \neg \tt{R} & \tt{Q}^{\tt{d}}  & \sqsubseteq  \neg \tt{R}^{\tt{d}} \\
\tt{p}(\tt{a}) & \leftarrow  \nf \tt{p}^{\tt{d}}(\tt{a}) & \tt{p}^{\tt{d}}(\tt{a})  & \leftarrow  \nf \tt{p}(\tt{a})\\
\tt{Q}(\tt{a}) & \leftarrow  & \tt{Q}^{\tt{d}}(\tt{a})  & \leftarrow  \nf \tt{NQ}(\tt{a})\\
\tt{R}(\tt{a}) & \leftarrow  \nf \tt{R}^{\tt{d}}(\tt{a}) & \tt{R}^{\tt{d}}(\tt{a})  & \leftarrow  \nf \tt{R}(\tt{a}), \nf \tt{NR}(\tt{a})
\end{align*}
Cf.\ the computation in Example~\ref{e:double}, $\tt{p}(\tt{a})$, $\tt{Q}(\tt{a})$, and $\tt{R}(\tt{a})$ are true in the sequence $\mathbf{P}_\omega^d$ while  $\tt{p}^{\tt{d}}(\tt{a})$, $\tt{Q}^{\tt{d}}(\tt{a})$, and $\tt{R}^{\tt{d}}(\tt{a})$ are false in the sequence $\mathbf{N}_\omega^d$.
The same results are derivable with a complete oracle.
$\tt{Q}(\tt{a})$ is derivable from the corresponding fact.
From that $\neg \tt{R}(\tt{a})$ is derivable and therefore $\nf \tt{R}^{\tt{d}}(\tt{a})$ as well.
This allows us to obtain $\tt{R}(\tt{a})$.
Now, $\tt{Q}(\tt{a})$ and $\tt{R}(\tt{a})$ together with $\cO$ are inconsistent from which we can derive $\tt{p}(\tt{a})$, but also $\neg \tt{Q}(\tt{a})$ and $\neg \tt{p}(\tt{a})$.
Consequently, $\nf \tt{p}^{\tt{d}}(\tt{a})$ and $\nf \tt{Q}^{\tt{d}}(\tt{a})$ hold as well, i.e., everything is supposedly true and false at the same time.

If we limit ourselves to the partial (consistent) oracle, then we no longer derive $\tt{p}(\tt{a})$, $\nf \tt{Q}(\tt{a})$, or $\nf \tt{p}(\tt{a})$.
In this case, $\tt{R}(\tt{a})$ is still true and false (inconsistent), but $\tt{Q}(\tt{a})$ is true, and $\tt{p}(\tt{a})$ is undefined.

Thus, the usage of such a partial oracle partially hides MKNF-inconsistencies and demonstrates a somewhat paraconsistent behavior instead. 
\end{example}
This example also shows why correctness of a partial oracle is only defined w.r.t.\ MKNF-consistent knowledge bases.
For MKNF-inconsistent knowledge bases the derivation relation is not identical in general.

By making assumptions on the complexity and number of results of an oracle, complexity results of \SLGO{} are obtained.

\begin{theorem}\label{t:dcpo}
Let $\mathcal{K}^d$ be a doubled Hybrid MKNF knowledge base, and $pT_{\cO}$ a correct partial oracle for $\cO$, such that for every goal $S$, the cardinality of $pT_{\cO}(I_{\cF},S,L)$ is bound by a polynomial on the number of program atoms.
Moreover, assume that computing each element of $pT_{\cO}$ is decidable with data complexity $\mathcal{C}$.
Then, the \SLGO{} evaluation of  a query in $\mathcal{K}^{d}$ is decidable with data complexity $\mathrm{P}^{\mathcal{C}}$.
\end{theorem}
\begin{proof}
Decidability is guaranteed by Theorem~\ref{t:terminator}.
As for complexity, first note that, given the polynomial data complexity of SLG \cite{CheW96}, \SLGO{} without calls to the oracle is of polynomial data complexity as well.
Considering the oracle, since the cardinality of $pT_{\cO}(I_{\cF_n},S,L)$ is bound by a polynomial, and each of the calls to the oracle can be seen as adding a new program rule (the result of  \orcr{} operation), only polynomially many such rules are added.
Hence, as such, the inclusion of oracle calls does not alter the the polynomial data complexity of SLG.
Now, computing each such rule amounts to a call to the oracle, which by hypothesis is decidable and with data complexity $\mathcal{C}$.
So, the overall data complexity is $\mathrm{P}^{\mathcal{C}}$.
Note that the doubling of the knowledge base does not affect that since the factor $2$ is subsumed by the (at least) polynomial complexity.
\end{proof}

In particular, Theorem~\ref{t:dcpo} means that if the partial oracle is tractable, and produces only polynomial many results, then \SLGO{} is also tractable.
Clearly, for an ontology part of the knowledge base that is tractable, it is possible to come up with a correct partial oracle that is also tractable.
Basically, all that needs to be done is to proceed with the usual entailment method, assuming that all program atoms hold, and collecting them for the oracle result.
To guarantee that the number of solutions of the oracle is bound by a polynomial, and still keeping with correctness, might be a bit more difficult. It amounts to finding a procedure that returns less results, and at the same time does not damage the completeness proof (similar to that of Theorem~\ref{th:correct}).
At least for the tractable case this is possible, albeit the oracle being the (polynomial complexity) bottom-up procedure that defines \MKNF.
This approach is, however, somewhat counterproductive to the whole idea of a top-down querying mechanism: we could simply use the bottom-up procedure in the first place to compute the model and store the results in a database which we then query on demand.
The following section defines a concrete oracle for the tractable description logic $\mathcal{EL}^{+}$ that maintains the desired data complexity and retains goal-orientation.

\section{An Oracle for $\mathcal{EL}^{+}$}\label{s:el+oracle}

When defining an oracle on $\mathcal{EL}^{+}$ we could simply try to
use the algorithm for subsumption presented in \cite{BBL05}: reduce
instance checking to subsumption and return the desired set of atoms
which, when proven, would ensure the derivability of the initial
query.  However, apart from the technical problems we would have to
face, like how to obtain these sets of atoms whose truth allows us to
prove the initial query, this would mean that we would have to run the
entire subsumption algorithm for each query posed to the oracle in
$\mathcal{EL}^{+}$.

We therefore proceed differently.  We still use the algorithm for
subsumption from \cite{BBL05} to compute the complete class hierarchy
of the $\mathcal{EL}^+$ TBox $\mathcal{T}$, but we use it only once, as a kind of
preprocessing of the ontology.  Then we take the obtained results
together with the TBox $\mathcal{T}$ and simplify them by removing all statements
that are redundant when looking for instances of classes in a top-down
manner.  The result, together with the $\mathcal{EL}^+$ ABox $\mathcal{A}$, is then turned into a set
of rules which can be used in a top-down manner, by using SLG alone,
yielding the desired oracle. Moreover, this way we can
straightforwardly combine these transformed rules with the ones in the
knowledge base and, with the top-down querying system defined in
Section \ref{s:SLGO}, obtain a single top-down procedure querying an
MKNF knowledge base where the ontology is described in
$\mathcal{EL}^{+}$.

\subsection{Subsumption in $\mathcal{EL}^+$}

In \cite{BBL05}, a polynomial time algorithm for subsumption was
described, and we recall important notions from it, restricted to
$\mathcal{EL}^{+}$.
For a TBox $\mathcal{T}$, the notion $\mathsf{BC}_{\mathcal{T}}$ represents the
smallest set of concept descriptions that contains all concept names
used in $\mathcal{T}$ plus the top concept $\top$; while
$\mathsf{R}_{\mathcal{T}}$ denotes the set of all role names used in
$\mathcal{T}$.  Using this notation, a normalized form of a TBox $\mathcal{T}$ is defined.

\begin{definition}\cite{BBL05} \label{def:normalForm}
A TBox $\mathcal{T}$ is in \emph{normal form} if 

\begin{enumerate}
\item all GCIs have one of the following forms, where $C_{1}$, $C_{2}\in \mathsf{BC}_{\mathcal{T}}$ and $D\in \mathsf{BC}_{\mathcal{T}}\cup \{\perp\}$:
\vspace{-.2in}

\[\begin{array}{lr@{\ }c@{\ }lp{.5in}lr@{\ }c@{\ }l}
(1) & C_{1}  & \sqsubseteq & D && (3) & \exists R.C_{1}  & \sqsubseteq & D\\
(2) & C_{1}\sqcap C_{2}  & \sqsubseteq & D && (4) & C_{1}  & \sqsubseteq & \exists R.C_{2} \\
\end{array}\]

\item all RI are of the form $R\sqsubseteq S$ or $R_{1} \circ R_{2} \sqsubseteq S$
\end{enumerate}
\end{definition}
By appropriately introducing new concept and role names, any TBox $\mathcal{T}$ can be turned into normal form and, as shown in \cite{BBL05}, this transformation can be done in linear time. So, from now on, we assume that any TBox $\mathcal{T}$ is in normal form.

The subsumption algorithm for $\mathcal{EL}^{+}$ (\cite{BBL05}) applies a set of completion rules to compute the entire class hierarchy, i.e. all subsumption relationships between all pairs of concept names occurring in $\mathcal{T}$.
In detail, given a normalized TBox $\mathcal{T}$, the algorithm computes:
\begin{itemize}
\item
a mapping $S$ from $\mathsf{BC}_{\mathcal{T}}$ to a subset of $\mathsf{BC}_{\mathcal{T}} \cup \{\perp\}$; and
\item
a mapping $T$ from $\mathsf{R}_{\mathcal{T}}$ to a binary relation on $\mathsf{BC}_{\mathcal{T}}$.
\end{itemize} 
These mappings make implicit relations explicit in the following sense:
\begin{description}
\item[(I1)] $D\in S(C)$ implies that $C\sqsubseteq D$, 
\item[(I2)] $(C,D)\in T(R)$ implies that $C\sqsubseteq \exists R.D$.
\end{description}
The initialization of these mappings is the following:
\begin{itemize}
\item $S(C) := \{C,\top\}$ for each $C\in \mathsf{BC}_{\mathcal{T}}$
\item $T(R) := \emptyset$ for each $R\in \mathsf{R}_{\mathcal{T}}$
\end{itemize}

Then the following completion rules are applied to extend $S$ and $T$ until no more rule applies.

\begin{itemize}
\item[CR1] If $C'\in S(C)$, $C'\sqsubseteq D\in \mathcal{T}$, and $D\not\in S(C)$ \\
then $S(C):=S(C)\cup \{D\}$
\item[CR2] If $C_1,C_2\in S(C)$, $C_1\sqcap C_2\sqsubseteq D\in \mathcal{T}$, and $D\not\in S(C)$ \\
then $S(C) := S(C)\cup \{D\}$
\item[CR3] If $C'\in S(C)$, $C'\sqsubseteq \exists R.D \in \mathcal{T}$, and $(C,D)\not\in T(R)$ \\
then $T(R) := T(R) \cup \{(C,D)\}$
\item[CR4] If $(C,D)\in T(R)$, $D'\in S(D)$, $\exists R.D' \sqsubseteq E\in \mathcal{T}$, and $E\not\in S(C)$ \\
then $S(C) := S(C) \cup \{E\}$
\item[CR5] If $(C,D)\in T(R)$, $\bot\in S(D)$, and $\bot\not\in S(C)$ \\
then $S(C) := S(C) \cup \{ \bot\}$
\item[CR6] If $(C,D)\in T(R)$, $R\sqsubseteq S\in \mathcal{T}$, and $(C,D)\not\in T(S)$ \\
then $T(S) := T(S) \cup \{(C,D)\}$
\item[CR7] If $(C,D)\in T(R_1)$, $(D,E)\in T(R_2)$, $R_1\circ R_2 \sqsubseteq R_3 \in \mathcal{T}$, and $(C,E)\not\in T(R_3)$ \\
then $T(R_3) := T(R_3)\cup \{(C,E)\}$
\end{itemize}
Note that we omitted the four completion rules related to nominals and concrete domains.

It is shown in \cite{BBL05} that this algorithm terminates in polynomial time and that it is correct.

\subsection{Simplifying the Ontology}

Given a normalized $\mathcal{EL}^{+}$ TBox $\mathcal{T}$
(Definition~\ref{def:normalForm}) and an $\mathcal{EL}^{+}$ ABox $\mathcal{A}$, the first step in transforming the ontology is to
apply the subsumption algorithm to $\mathcal{T}$ and obtain the
mappings $S$ and $T$ computed by it.  In particular, we obtain via $S$ all
the subsumption relationships implicitly or explicitly present in
$\mathcal{C}$.
In fact, it is easy to see that the initialization of $C\in S(C)$ for
each $C\in \mathsf{BC}_{\mathcal{T}}$ ensures that each GCI of the
form (1) of the normal form of Definition~\ref{def:normalForm}
($C_{1} \sqsubseteq D$) is also obtained by $D\in S(C_{1})$, and each
GCI of the form (4) ($ C_{1} \sqsubseteq \exists R.C_{2} $) is
obtained by $(C_{1},C_{2})\in T(R)$.\footnote{Cf.\ the completion rules
$CR1$ and $CR3$ in Section \ref{s:prel} which precisely add each such
explicit GCIs to the appropriate mapping.}

It follows immediately from that, that we can actually ignore all GCIs of the form (1) and (4) as long as we have the complete mappings $S$ and $T$ of the subsumption algorithm available.
But we can simplify even more.

\begin{example}
Consider the Hybrid MKNF knowledge base with $\cO$ in $\mathcal{EL}^+$, containing one rule, and some facts.
\[\begin{array}{r@{\ }c@{\ }lp{.5in}l}
\tt{C} & \sqsubseteq & \exists \,\tt{R}.\tt{D} && \tt{G}(\tt{x})  \leftarrow  \tt{D}(\tt{x}) \\
\exists \,\tt{R}.\tt{C} & \sqsubseteq & \tt{D}  && \tt{C}(\tt{a}).\ \ \ \ \ \  \tt{C}(\tt{b}).\\
\tt{C}_{1} \sqcap \tt{C}_{2} & \sqsubseteq & \tt{D} && \tt{R}(\tt{a},\tt{b}). 
\end{array}\]
Now consider that we want to know whether $\tt{G}(\tt{a})$ holds.
There is only one rule that allows us to derive $\tt{G}(\tt{a})$, and this requires that $\tt{D}(\tt{a})$ is derivable.
Obviously, if we have $\tt{C}_{1}(\tt{a})$ and $\tt{C}_{2}(\tt{a})$ then $\tt{D}(\tt{a})$ holds as well.
But this information is currently not present in the knowledge base.
If we check the second GCI then obtaining $\tt{D}(\tt{a})$ requires finding $\tt{R}(\tt{a},\tt{x})$ and $\tt{C}(\tt{x})$ which appear as facts in the rule part, for $\tt{x}=\tt{b}$.
Intuitively, we want the oracle to transform the query $\tt{D}(\tt{a})$ into an \SLGO{} node $\tt{D}(\tt{a}):- \mid \tt{R}(\tt{a},\tt{x}), \tt{C}(\tt{x})$, the goals of which can then be resolved, leading to a derivation of $\tt{D}(\tt{a})$.

Next, suppose we alternatively query for $\tt{G}(\tt{b})$, and subsequently query the oracle for $\tt{D}(\tt{b})$.
Then the second GCI does not allow us to derive $\tt{D}(\tt{b})$ because there is no $\tt{R}(\tt{b},\tt{x})$ for some $\tt{x}$ derivable; the third does not allow us to derive $\tt{D}(\tt{b})$ because there are no individuals known to hold in $\tt{C}_{1}$ or $\tt{C}_{2}$.
But even using the first GCI does not allow us to derive $\tt{D}(\tt{b})$: while $\tt{C}(\tt{a})$ holds and we know that there is an explicit relation $\tt{R}(\tt{a},\tt{b})$ in the knowledge base, the semantics of $\cO$ (and descriptive first-order semantics in general) does not allow to derive $\tt{D}(\tt{b})$, since $\tt{D}(\tt{b})$ does not hold in all models of $\cO$ - there are models where $\tt{R}(\tt{a},\tt{i})$ and $\tt{D}(\tt{i})$ hold for some individual $\tt{i}$ not appearing in the knowledge base. 
\end{example}
Clearly in a $\mathcal{EL}^+$ KB with a normalized TBox $\mathcal{T}$, GCIs of the form (3) ($\exists R.C_{1} \sqsubseteq D$) and (2) ($C_{1}\sqcap C_{2} \sqsubseteq D$) -- and therefore also of the form (1) -- can be used to derive information when answering an (instance) query.
On the other hand, the example implies that GCIs of the form (4) ($C_{1} \sqsubseteq \exists R.C_{2}$) do not contribute to drawing this kind of conclusions.
We now formalize this observation.

For simplicity of notation, we start by transforming all the mappings obtained from the algorithm into GCIs, and then we remove all GCIs of the form (4).

\begin{definition}
Let $\mathcal{T}$ be an $\mathcal{EL}^{+}$ TBox and $S$ and $T$ be the mappings obtained from the subsumption algorithm.
We obtain the \emph{completed $\mathcal{EL}^{+}$ TBox} $\mathcal{T}'$ from $\mathcal{T}$ by adding for each  $D\in S(C)$ a GCI $C\sqsubseteq D$ to $\mathcal{T}'$ and for each $(C,D)\in T(R)$ a GCI $C\sqsubseteq \exists R.D$ to $\mathcal{T}'$.

Let $\mathcal{T}$ be a completed $\mathcal{EL}^{+}$ TBox.
We define the \emph{reduced $\mathcal{EL}^{+}$ TBox} $\mathcal{T}'$ which is obtained from the completed TBox $\mathcal{T}$ by removing all GCIs of form (4).
\end{definition}
\noindent
It is straightforward to see that the transformation from the TBox $\mathcal{T}$ to the completed TBox $\mathcal{T}'$ simply allows us to disregard the mappings $S$ and $T$ obtained by the algorithm of subsumption without losing any of the subset relationships contained in these mappings.

Now we have to show that a reduced TBox, which in general does not preserve $\mathcal{EL}^{+}$ semantics, is still suitable for the query answering we are interested in, which restricts itself to queries of the form $C(a)$ or $R(a,b)$.

\begin{proposition}\label{p:equiCBox}
Let $\mathcal{A}$ be an $\mathcal{EL}^{+}$ ABox, $\mathcal{T}$ be a completed $\mathcal{EL}^{+}$ TBox, and $\mathcal{T}'$ the reduced $\mathcal{EL}^{+}$ TBox obtained from $\mathcal{T}$.
Then the following two conditions hold.
\begin{itemize}
\item[(i)] $a$ is an instance of concept $C$ in $\mathcal{A}$ w.r.t. $\mathcal{T}$ iff $a$ is an instance of concept $C$ in $\mathcal{A}$ w.r.t. $\mathcal{T}'$.
\item[(ii)] $(a,b)$ is an instance of role $R$ in $\mathcal{A}$ w.r.t. $\mathcal{T}$ iff $(a,b)$ is an instance of role $R$ in $\mathcal{A}$ w.r.t. $\mathcal{T}'$.
\end{itemize}
\end{proposition}
\begin{proof}
For (i) we have to show that $a^{\mathcal{I}}\in C^{\mathcal{I}}$ for every common model $\mathcal{I}$ of $\mathcal{A}$ and $\mathcal{T}$ iff $a^{\mathcal{I}}\in C^{\mathcal{I}}$ for every common model $\mathcal{I}$ of $\mathcal{A}$ and $\mathcal{T}'$; for (ii) we have to show that
 $(a^{\mathcal{I}},b^{\mathcal{I}})\in R^{\mathcal{I}}$ for every common model $\mathcal{I}$ of $\mathcal{A}$ and $\mathcal{T}$ iff $(a^{\mathcal{I}},b^{\mathcal{I}})\in R^{\mathcal{I}}$ for every common model $\mathcal{I}$ of $\mathcal{A}$ and $\mathcal{T}'$.
 We are going to sketch the argument for (i); the case of (ii) follows analogously.
 
\noindent $'\Leftarrow'$: follows directly from monotonicity: adding GCIs of the form (4) will not invalidate any drawn conclusions, i.e. if $a^{\mathcal{I}}\in C^{\mathcal{I}}$ for every common model $\mathcal{I}$ of $\mathcal{A}$ and $\mathcal{T}'$ then adding GCIs of the form (4) can only reduce the common models of $\mathcal{I}$ of $\mathcal{A}$ and never increase. We conclude $a^{\mathcal{I}}\in C^{\mathcal{I}}$ for every common model $\mathcal{I}$ of $\mathcal{A}$ and $\mathcal{T}$.

\noindent $'\Rightarrow'$: suppose that $a^{\mathcal{I}}\in C^{\mathcal{I}}$ for every common model $\mathcal{I}$ of $\mathcal{A}$ and $\mathcal{T}$.
If none of the GCIs of the form (4) contains the concept name $C$ then we can remove them all and $a^{\mathcal{I}}\in C^{\mathcal{I}}$ for every common model $\mathcal{I}$ of $\mathcal{A}$ and $\mathcal{T}'$.
The same argument applies if $C$ appears only on the left hand side of such GCIs.
So assume $C$ appears on the right hand side of at least one such GCI $C_{1} \sqsubseteq \exists R.C$.
However, even if there is an individual $i$ such that $i^{\mathcal{I}}\in C_{1}^{\mathcal{I}}$ and $(i^{\mathcal{I}},a^{\mathcal{I}})\in R^{\mathcal{I}}$ for every common model $\mathcal{I}$ of $\mathcal{A}$ then $\mathcal{T}$ does not allow to conclude $a^{\mathcal{I}}\in C^{\mathcal{I}}$ for every common model $\mathcal{I}$ of $\mathcal{A}$ and $\mathcal{T}$.
We can thus conclude that $a^{\mathcal{I}}\in C^{\mathcal{I}}$ for every common model $\mathcal{I}$ of $\mathcal{A}$ and $\mathcal{T}'$.
\end{proof}

Having proven that  
TBox completion does not alter the derivability
of instance queries, we can take a short cut: instead of completing
the TBox we can directly remove all GCIs of the form (4) and discard
the mapping $T$.  We then complete the TBox only with respect to the
mapping $S$ and obtain the reduced TBox.

\begin{corollary}
Let $\mathcal{T}$ be a $\mathcal{EL}^{+}$ TBox and $S$ and $T$ be the mappings obtained from the subsumption algorithm.
We obtain the reduced TBox $\mathcal{T}'$ from $\mathcal{T}$ by removing all GCIs of the form (4) from $\mathcal{T}$ and by adding for each  $D\in S(C)$ a GCI $C\sqsubseteq D$.
\end{corollary}

\subsection{Transformation into Rules}

Now, we show how to transform the reduced $\mathcal{EL}^{+}$ KB into rules in such a way that running the SLG procedure on the obtained set of rules yields an oracle that can be used in \SLGO.
Special care must be taken with inconsistencies and with the fact that if an atom is proven false in the ontology, then its negation also holds in the rules.
Note that this is achieved in \SLGO{} by querying for classically negated atoms, but these are outside
the syntax of $\mathcal{REL}$ even though a restricted form of negation is achievable via $\bot$.

Regarding inconsistencies, there are two kinds which can appear in the three-valued Hybrid MKNF semantics as presented in \cite{KAH:AI11}: either the ontology alone is inconsistent, or there is an inconsistency resulting from contradictory derivations in the rules and the ontology.
In the first case, there is not much to be done.
An inconsistent ontology has no models and we can simply derive anything from it, making reasoning over a combined knowledge base rather pointless.
We therefore admit in the following an a-priori consistency check of the ontology alone, and proceed only if it succeeds, i.e., we limit ourselves in the following to a consistent ontology.\footnote{Note that ontologies in $\mathcal{EL}^{+}$ can in fact be inconsistent: consider a GCI $C\sqsubseteq \perp$ in the TBox and an assertion $C(a)$ in the ABox.}
For the second case, the bottom-up computation allows us to detect such problems, but in \SLGO{} we are limited to finding atoms that are true and false at the same time, i.e., if for some $\tt{C}(\tt{a})$ both queries $\tt{C}(\tt{a})$ and $\nf \tt{C}^{\tt{d}}(\tt{a})$\footnote{Recall that we use the doubled predicate for determining falsity.} are answered with 'yes', then the combined KB is inconsistent.
This can, of course, not be complete for a partial oracle, as shown in Example~\ref{e:double2}, so that we obtain a paraconsistent behavior.
To carry over this behavior to a transformation into rules, we have to take into consideration the transformation presented in Definition \ref{d:MKNF-cohtransformKd} and their effect on the $\mathcal{EL}^+$ KB.

Regarding classical negation, we solve the problem in a specific way.
In \SLGO, the special negative literals $\nf NH(t_i)$ are used to call $\neg H(t_i)$.
Since this is not expressible in $\mathcal{EL}^+$ we simply consider $\nf NH(t_i)$ as normal negative literals, and transform $\cO$ into rules such that $\nf NH(t_i)$ holds if $\neg H(t_i)$ holds.
More precisely, if $H\sqsubseteq \bot$, then $NH(t_i)$ holds.

We are now ready to define the transformation of the ontology $\cO$ consisting of a reduced $TBox$ and an ABox into a set of already doubled rules (see Definition \ref{d:doubling}).

\begin{definition}\label{d:transfcons}
Let $\mathcal{K}=(\mathcal{O},\mathcal{P})$ be a Hybrid MKNF knowledge base with a consistent $\mathcal{EL}^{+}$  KB $\cO$.
We define $\mathcal{P}_{\mathcal{O}}^{d}$ from $\cO$, where $C$,$D$, $C_{1}$, and $C_{2}$ are concept names, $R$, $S$, $T$ are role names, and $a$, $b$ are individual names, as the smallest set containing:
\begin{description}
\item[(a1)] for each $C(a)\in \cA$: $C(a)\leftarrow$ and $C^{d}(a)\leftarrow \nf NC(a)$.
\item[(a2)] for each $R(a,b)\in \cA$: $R(a,b)\leftarrow$ and $R^{d}(a,b)\leftarrow \nf NR(a,b)$.
\item[(c1)] for each GCI $C\sqsubseteq D \in \mathcal{T}$: $D(x) \leftarrow C(x)$ and \\ $D^{d}(x) \leftarrow C^{d}(x), \nf ND(x)$.
\item[(c2)] for each $C_{1}\sqcap C_{2} \sqsubseteq D\in \mathcal{T}$: $D(x) \leftarrow C_{1}(x),C_{2}(x)$ and \\ $D^{d}(x) \leftarrow C_{1}^{d}(x),C_{2}^{d}(x),\nf ND(x)$.
\item[(c3)] for each $\exists R.C \sqsubseteq D\in \mathcal{T}$: $D(x) \leftarrow R(x,y), C(y)$ and \\ $D^{d}(x) \leftarrow R^{d}(x,y), C^{d}(y),\nf ND(x)$.
\item[(r1)] for each RI $R\sqsubseteq S \in \mathcal{T}$: $S(x,y) \leftarrow R(x,y)$ and \\ $S^{d}(x,y) \leftarrow R^{d}(x,y), \nf NS(x,y)$.
\item[(r2)] for each $R\circ S \sqsubseteq T\in \mathcal{T}$: $T(x,z) \leftarrow R(x,y),S(y,z)$ and \\ $T^{d}(x,z) \leftarrow R^{d}(x,y),S^{d}(y,z), \nf NT(x,z)$.
\item[(i1)] for each $C\sqsubseteq \perp \in \mathcal{T}$: $NC(x) \leftarrow$.
\item[(i2)] for each $C_{1}\sqcap C_{2} \sqsubseteq \perp\in \mathcal{T}$: $NC_{2}(x) \leftarrow C_{1}(x)$ and $NC_{1}(x) \leftarrow C_{2}(x)$.
\item[(i3)] for each $\exists R.C \sqsubseteq \perp\in \mathcal{T}$: $NC(y) \leftarrow R(x,y)$ and $NR(x,y) \leftarrow C(y)$ .
\end{description}
\end{definition}
Note that the cases (i1) to (i3) are used to introduce truth of some $NH(t_i)$.
Furthermore, these three cases only produce one rule, since atoms based on predicates of the forms $NC^d$ or $NR^d$ are not required anywhere.

Program $\mathcal{P}_{\mathcal{O}}^{d}$ can then be used as the basis for obtaining a correct partial oracle for   $\mathcal{EL}^{+}$, to be integrated in the general procedure of \SLGO. 
Recall that an oracle receives a query $S$ and the already derived (positive) information $I_{\mathcal{F}_n}^+$, and returns a set of atoms $L$, which if proven, ensure that $S$ is derivable.
The general idea of such an oracle for $\mathcal{EL}^{+}$ would be to use SLG to query $Q$ in a program consisting of  $\mathcal{P}_{\mathcal{O}}^{d}$ plus facts for all the atoms in $I_{\mathcal{F}_n}^+$,  in such a way that any time an atom also defined in the rules is queried, the atom can succeed, i.e., is removed from the resolvent, and is collected in a set associated to the respective derivation branch.\footnote{An alternative way of viewing this, would be to add to $\mathcal{P}_{\mathcal{O}}^{d}$ facts for all the atoms defined in the rules, run SLG as usual, but collecting all those facts that were used in the derivation.}
Upon success, the so modified SLG procedure would return the set of collected atoms.
The partial oracle would be defined by the relation with the query, the running forest, and the returned set of collected atoms.
However, since both the rule part and the oracle itself would be evaluated by an SLG procedure, they can be combined: instead of collecting the atoms in the set, and then calling them in \SLGO{} after the oracle returns a result, one can simply immediately call the otherwise collected atoms, i.e., the atoms defined in the program.
This way, correctness of the so defined partial oracle is equivalent to the correctness of the above transformation.
We start by proving this for the consistent case:

\begin{theorem}
Let $\mathcal{K}=(\mathcal{O},\mathcal{P})$ be an MKNF-consistent Hybrid MKNF knowledge base with $\cO$ in $\mathcal{EL}^{+}$.
Then $\mathcal{K}_{\mathcal{EL}^{+}}=(\emptyset, (\mathcal{P}^{d}\cup \mathcal{P}_{\mathcal{O}}^{d}))$ is semantically equivalent to $\mathcal{K}^{d}=(\cO,\mathcal{O}^{d},\mathcal{P}^{d})$.
\end{theorem}

\begin{proof}
We have to show that $\mathcal{P}_{\mathcal{O}}^{d}$ is equivalent to $\cO$ and $\mathcal{O}^{d}$.

The transformations on ABox assertions, (a1) and (a2), on GCIs in $\cC$, (c1), (c2), and (c3), and on role inclusions, (r1) and (r2), are semantically equivalent and can be found, e.g., in \cite{GHVD03}.
Since $\mathcal{O}$ contains the original GCIs and $\mathcal{O}^d$ the doubled ones with new predicate names, we also create two rules, one for each of the two DL knowledge bases in $\mathcal{K}^d$.
Note that the addition of predicates, such as $NC(x)$, to the body of a rule with head $C^{d}(x)$ is just done to enforce that whenever $NC(x)$ holds, i.e., $\neg C(x)$, then $C^{d}(x)$ cannot become true, which is used in the consistent case to enforce coherence. 
We only have to consider the transformations (i1) to (i3).
\begin{itemize}
\item[(i1)] $C\sqsubseteq \perp$: $C$ is unsatisfiable, i.e., $\neg C(x)$ for all $x$;
$\mathcal{O}$ contains a statement that allows us to infer $\neg C(x)$ which corresponds exactly to the fact $NC(x)\leftarrow$.
\item[(i2)] $C_{1}\sqcap C_{2} \sqsubseteq \perp$: the statement expresses disjointness of $C_{1}$ and $C_{2}$, i.e., $\neg (C_{1}(x)\wedge C_{2}(x))$ for all $x$ which is equivalent to $C_{1}(x) \rightarrow \neg C_{2}(x)$ and $C_{2}(x) \rightarrow \neg C_{1}(x)$; using the correspondences $\neg C_{1}(x) \rightarrow NC_{1}(x)$ and $\neg C_{2}(x) \rightarrow NC_{2}(x)$.
\item[(i3)] $\exists R.C \sqsubseteq \perp$ follows the same argument as (i2).
\end{itemize}
This finishes the proof.
\end{proof}

For the MKNF-inconsistent case, we point out that one result of the
transformation into rules is that we obtain a somewhat paraconsistent
approach: while an inconsistent ontology allows us to derive anything
from it, the process of doubling the rules enables us to derive those
consequences that do not depend on inconsistent information contained
in the KB as presented in Example~\ref{e:double2}.  We leave further
details of this paraconsistency to future studies.

Finally, we have to show that the process of translating the ontology into rules and reasoning over the combined set of rules with \SLGO{} also preserves the intended polynomial data complexity.

\begin{theorem}
Let $\mathcal{K}=(\mathcal{O},\mathcal{P})$ be a Hybrid MKNF knowledge base with $\cO$  in $\mathcal{EL}^{+}$.
An \SLGO{} evaluation of a query in $\mathcal{K}_{\mathcal{EL}^{+}}=(\emptyset, (\mathcal{P}^{d}\cup \mathcal{P}_{\mathcal{O}}^{d}))$ is decidable with data complexity in $\mathrm{P}$.
\end{theorem}
\begin{proof}
The $\mathcal{EL}^+$ oracle is in fact a transformation of the
evaluation of the $\mathcal{EL}^{+}$ ontology into a set of rules so
that evaluation of the Hybrid MKNF KB is made w.r.t.\ a
combined set of rules. Note that the polynomial subsumption
algorithm for $\mathcal{EL}^{+}$ and the linear transformations to
obtain $\mathcal{K}_{\mathcal{EL}^{+}}$ together are in $\mathrm{P}$.

We consider the data complexity to be the number
of answers returned for a given atomic query w.r.t.\ the number
ground facts in the rules, and the number of assertions in the ABox.
Note that the transformation of the
$\mathcal{EL}^{+}$ axioms introduces a number of facts in the
rules at most linear in the size of the ABox (cases
{\em a1} and {\em a2} of Definition~\ref{d:transfcons}).  As a result
the number of facts in the transformed system will be linear in the
size of the rule facts plus the ABox of the original system.

Finally, note that Theorem \ref{t:dcpo} ensures polynomial data
complexity of query evaluation in Hybrid MKNF.
The transformed KB $\mathcal{K}_{\mathcal{EL}^{+}}$ can be considered 
a Hybrid MKNF KB with empty $\cO$, so that by Theorem \ref{t:dcpo} the transformed 
KB has a data complexity in $\mathrm{P}$.  
Since the transformed KB increases the size of the rule facts linearly this proves the
statement.
\end{proof}

\section{Discussion and Conclusions}\label{s:con}

\subsection{Related Work}\label{ss:relwork}
Three other semantics define well-founded models for a combination of
rules and ontologies, namely the works in \cite{EILS11}, \cite{Luk10},
and \cite{DM07}.  The approach of \cite{EILS11} combines ontologies
and rules in a modular way, keeping separate the 
semantics of both, and has identical data complexity to the
well-founded MKNF semantics for a tractable DL.  As such, it has
similarities with \SLGO{} in terms of reasoning, in the sense that
both treat reasoning in the DL separately.
However, the approach of \cite{EILS11}, implemented using the dlv hex
system \cite{EIST06}, has a looser integration, limiting the way the
ontology can call back program atoms.  In \cite{EILS11}, the set of
atoms occurring in rules and DLs are disjoint, and links must be
established using specific interface atoms in the rules, which can
only temporarily add information to the DL part.  To the contrary, our
semantics does not require any such restriction so that the flow of
information between rules and DLs is not limited.
The well-founded semantics for normal dl-programs \cite{Luk10} does
not require any of these limitations either, but it requires that the
ontology is decomposable into a positive and a negative part.  This
severely restricts the applicability to arbitrary DLs although it is
shown in \cite{Luk10} that the approach is applicable to the $DL$-$Lite$
family.  This contrasts with \SLGO{}, which can be applied to
any decidable DL.
Hybrid programs of \cite{DM07} are even more restrictive
than \cite{EILS11} in the combination: in fact it only allows to
transfer information from the ontology to the rules and not the other
way around.  Moreover, the semantics of this approach differs from
MKNF \cite{MR10,KAH:AI11} and also \cite{EILS11,Luk10} in that if an
ontology expresses $B_{1} \vee B_{2}$, then the semantics
in \cite{DM07} derives $p$ from rules $p\leftarrow B_{1}$ and
$p\leftarrow B_{2}$, $p$ while MKNF and \cite{EILS11,Luk10} do not.  More
generally, several well-founded models may exist, contrary to the more
common definitions of well-founded models.

In Section~\ref{s:el+oracle}, we also presented a concrete oracle
for \SLGO{} that allows the combination of non-monotonic rules with
the DL $\mathcal{EL}^{+}$.  Using this oracle, \SLGO{} remains
tractable w.r.t.\ data complexity, and permits the discovery of
possible inconsistencies between the rules and the ontology.
These results contribute to the work related to conjunctive query
answering with respect to $\mathcal{EL}^{+}$.  Conjunctive query
answering has been studied, e.g., for acyclic $\mathcal{EL}^{+}$
in \cite{MLXKF09} as an extension to \cite{LTW09}, where the
limitation to acyclic TBoxes avoids general undecidability
(see \cite{Rosa07c}).  In contrast, our work limits the queries to be
DL-safe but adds rules as an additional expressive means.  As an
additional point of comparison, since the concrete oracle operates in
a kind of abductive way -- by finding the set of atoms which together
with the ontology prove the query - our work also bears some relation
to \cite{bienvenu-kr08} where general complexity results on abduction
for $\mathcal{EL}^{++}$ are established.
Another concrete oracle for \SLGO{} was very recently presented
in \cite{KA:ISWC11} providing a top-down procedure for
$DL$-$Lite_\mathcal{R}$, the DL underlying
the tractable OWL 2 profile, OWL 2 QL.  As does the $\mathcal{EL}^{+}$
oracle, the $DL$-$Lite_\mathcal{R}$ oracle maintains the data
complexity of the bottom-up approach.

\subsection{Conclusions}
Together with the alternate computation method of
Section~\ref{s:altcomp}, \SLGO{} provides a sound and complete
querying method for Hybrid MKNF knowledge bases.  
Further, \SLGO{} maintains the favorable computational complexity of
the well-founded MKNF model and freely allows bidirectional calls
between the ontology and the rules, unlike other approaches (as
discussed in Section \ref{ss:relwork}).  As such it presents a
significant step towards making Hybrid MKNF knowledge bases
practically usable for the Semantic Web.

Future work with regard to concrete oracles includes the (non-trivial)
extension of the $\mathcal{EL}^+$ oracle to $\mathcal{EL}^{++}$.  A
second potentially fruitful extension is the construction of an oracle
for ELP (\cite{KRH:ELP-08}), an approach based on rules that allow DL
expressions instead of (negated) atoms and that covers
$\mathcal{EL}^{++}$.  Since the altorithmization of ELP, like that of
the $\mathcal{EL}^+$ oracle, transforms its expressive rules into
datalog rules it may benefit from the pre-processing step introduced
for $\mathcal{EL}^+$ knowledge bases.  A third concrete oracle to
investigate is $\mathcal{SROELV}_n$ \cite{elp2}, a tractable fragment
of $\mathcal{SROIQ}$ enhanced with nominal schemas that covers not
only datalog rules in DL syntax but also $\mathcal{EL}^{++}$.
 
Other future work will address the class of conjunctive queries
that \SLGO{} can answer.  While \SLGO{} queries posed to KBs without
an ontology are handled in the same way as in SLG, the queries posed
to the ontology, which are required to be ground, are not conjunctive
queries in the sense of \cite{GLHS08}, where boolean queries may
contain anonymous variables that are interpreted existentially. 
The extension to such queries may possibly by supported by anonymous
variables in XSB, the system in which \SLGO{} is currently
implemented.

Furthermore, we may take evolution and dynamics into consideration.
In \cite{SlotaICLP10,SlotaICLP11}, updating Hybrid MKNF knowledge
bases is considered, while \cite{SlotaECAI10,SlotaLPNMR11} presents
the problem from a more general perspective in SE-models.  The
extension of \SLGO{} to such dynamic knowledge bases forms another
line of future work.

Finally, we mention that a prototype implementation of \SLGJMT{}
exists \cite{GAS10:padl} based on XSB Prolog and its ontology
management library CDF.  Because CDF includes an ${\cal ALCQ}$ prover
written directly using XSB, the \orcr{} operation of
Section~\ref{s:SLGO} is more easily implemented than it would be using
a separate prover, as is the detection of when a mutually dependent
set of subgoals is completely evaluated
(Definition~\ref{def:comp-eval}).  Accordingly, the polynomial data
complexity of the oracle is also more easily guaranteed.  The
resulting implementation will enable further study into how Hybrid
MKNF knowledge bases can be practically used and will indicate needed
optimizations and useful extensions. For instance, since XSB supports
constraint processing, temporal or spatial constraints can be added to
the ABox.  From a systems perspective, the multi-threading of XSB can
allow for the construction of Hybrid MKNF knowledge servers that make
use of either Prolog rules or F-logic rules (via FLORA-2, which is
implemented using XSB). As mentioned in Section~\ref{s:prop} the final
forest of a \SLGO{} evaluation produces a well-founded reduct of the
rules and oracle information.  This reduct, which is materialized in
XSB's tables, can be sent to a stable model generator through XSB's
XASP library to obtain a partial stable MKNF model of \cite{MR10}.

\bibliography{slgO}
\bibliographystyle{acmtrans}


\end{document}